\pdfoutput=1
\documentclass[11pt]{article}
\usepackage[preprint]{acl}
\usepackage{times}
\usepackage{latexsym}
\usepackage[T1]{fontenc}
\usepackage[utf8]{inputenc}
\usepackage{microtype}
\usepackage{inconsolata}
\usepackage{graphicx}
\usepackage{booktabs}
\usepackage{amsmath}
\usepackage{amssymb}
\usepackage{xcolor}
\usepackage{pifont}
\usepackage{tikz}
\usetikzlibrary{arrows.meta,patterns,positioning,calc}
\usepackage{caption}
\newcommand{\sF}{\textsc{S-F}}
\newcommand{\sC}{\textsc{S-C}}
\newcommand{\jfc}{\textsc{J-FC}}
\newcommand{\jcf}{\textsc{J-CF}}

\title{Do the Details Help the Whole? One-Way Transfer from Person Actions\\to Group Intent in VLMs on JRDB-Social}

\author{
  \textbf{Chengguang Gan}\textsuperscript{1},
  \textbf{Zhixi Cai}\textsuperscript{2},
  \textbf{Yunhao Liang}\textsuperscript{3},
  \textbf{Hanjun Wei}\textsuperscript{3},
\\
  \textbf{QingHao Zhang}\textsuperscript{4},
  \textbf{Shiwen Ni}\textsuperscript{5}
\\
  \textsuperscript{1}Independent Researcher \quad
  \textsuperscript{2}Monash University \quad
  \textsuperscript{3}University of Chinese Academy of Sciences
\\
  \textsuperscript{4}Pusan National University \quad
  \textsuperscript{5}Shenzhen University of Advanced Technology
\\
  \small{\texttt{ganchengguan@yahoo.co.jp}, \texttt{zhixi.cai@monash.edu}, \texttt{liangyunhao22@mails.ucas.ac.cn},}
\\
  \small{\texttt{weihanjun23@mails.ucas.ac.cn}, \texttt{zhangqinghao@pusan.ac.kr}, \texttt{nishiwen@suat-sz.edu.cn}}
}

\begin{document}
\maketitle
\begin{abstract}
Social scene datasets label people at two levels: what each person is doing and what the group is there for. Training on both levels at once is often assumed to help each of them. We test this assumption for vision-language models (VLMs) on JRDB-Social, treating person actions as the details and group intent as the whole. We fine-tune Qwen3.5 4B under eight training arms with three seeds, repeat the core arms on Qwen3.5 9B and Qwen3-VL 4B, and judge every arm with a verdict rule fixed before testing. Writing each member's actions before the intent raises intent macro-F1 from 30.1 to 31.0, and writing the intent first reaches 31.2, while action F1 stays within 0.2 points of an action-only model. Under the selected recipe, no arm trained with intent improves actions. The gain requires correct, image-grounded actions: empty action lists or actions borrowed from another group with the same intent lower intent F1. Forced-prefix tests, occlusion, member masking, linear probes and component grafting show that the intent follows the written actions, relies more on people and partly resides in the vision tower. Supplying actions as input instead of as targets hurts both levels.
\end{abstract}

\section{Introduction}
\label{sec:intro}

A robot that moves through a station or a campus has to read people at two levels \citep{mavrogiannis2023core}. It needs to know what each person is doing, such as standing, walking or looking at the robot, and it needs to know what a group is there for, such as waiting for a bus or heading to class. Datasets recorded from robots annotate both levels \citep{martin2021jrdb,jahangard2024jrdb,ehsanpour2022jrdb,han2022panoramic}, and models for group activity have long assumed that the person level supports the group level \citep{ibrahim2016hierarchical,ehsanpour2020joint}. Several recent systems go further and describe the two levels as mutually reinforcing \citep{lee2024spatio,gan2024mpt}. In text, sentence-level and word-level labels have been reported to reinforce each other in the same way \citep{gan2023sentence,gan2026multilingual}.

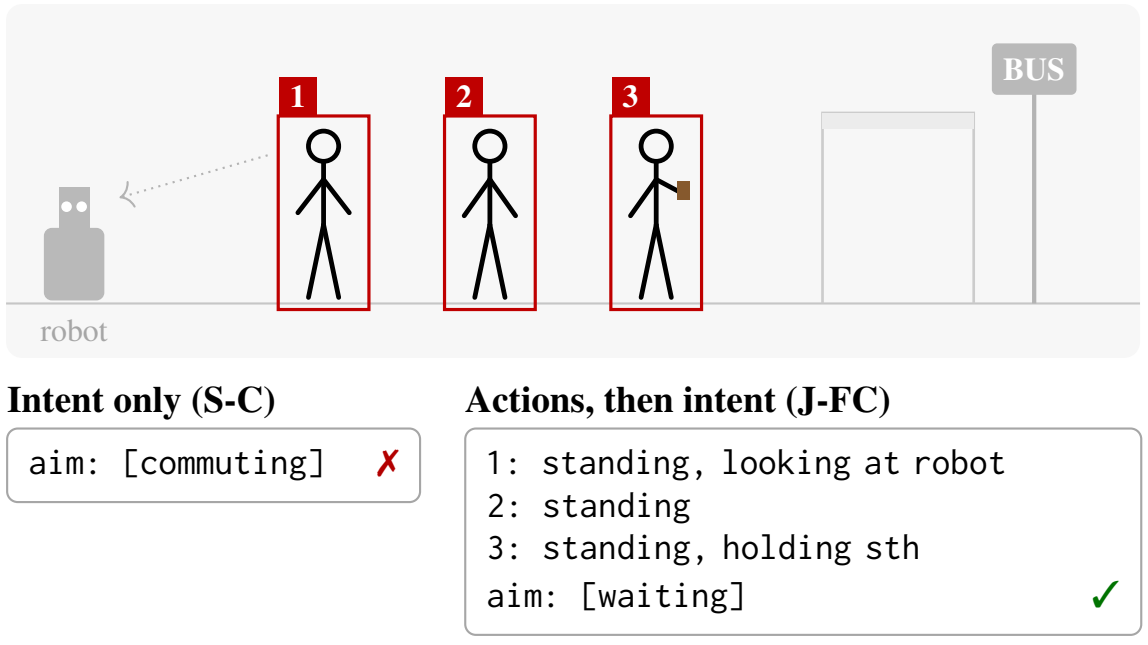
\begin{figure}[t]
\centering
\begin{tikzpicture}[
  font=\scriptsize, x=1cm, y=1cm,
  person/.style={line width=0.9pt, line cap=round},
  box/.style={draw=red!75!black, line width=0.6pt},
  tag/.style={fill=red!75!black, text=white, font=\tiny\bfseries, inner sep=1pt, minimum size=7pt},
  outb/.style={draw=black!35, rounded corners=2pt, fill=white, align=left, inner sep=4pt, font=\scriptsize\ttfamily},
  hdr/.style={font=\scriptsize\bfseries, anchor=north west, inner sep=0pt},
]
\fill[gray!6, rounded corners=3pt] (-3.75,-1.12) rectangle (3.75,1.22);
\draw[gray!45, line width=0.4pt] (-3.75,-0.76) -- (3.75,-0.76);          
\draw[gray!60, line width=0.8pt] (3.05,-0.76) -- (3.05,0.70);
\fill[gray!55, rounded corners=1pt] (2.77,0.62) rectangle (3.33,0.96);
\node[font=\tiny\bfseries, text=white] at (3.05,0.79) {BUS};
\draw[gray!40, line width=0.5pt] (1.65,-0.76) -- (1.65,0.50) -- (2.65,0.50) -- (2.65,-0.76);
\fill[gray!15] (1.65,0.40) rectangle (2.65,0.50);
\fill[gray!55, rounded corners=1.5pt] (-3.50,-0.74) rectangle (-3.10,-0.26);
\fill[gray!55] (-3.40,-0.26) rectangle (-3.20,0.01);
\fill[white] (-3.35,-0.12) circle (0.035); \fill[white] (-3.25,-0.12) circle (0.035);
\node[font=\tiny, text=gray!70] at (-3.30,-0.94) {robot};
\newcommand{\figperson}[2]{%
  \draw[person] (#1,0.28) circle (0.10);
  \draw[person] (#1,0.17) -- (#1,-0.24);
  \draw[person] (#1,-0.24) -- (#1-0.10,-0.72);
  \draw[person] (#1,-0.24) -- (#1+0.10,-0.72);
  #2}
\figperson{-1.65}{\draw[person] (-1.65,0.06) -- (-1.82,-0.16); \draw[person] (-1.65,0.06) -- (-1.48,-0.16);
  \draw[->, gray!60, line width=0.4pt, densely dotted] (-2.02,0.22) -- (-3.00,-0.06);}
\figperson{-0.55}{\draw[person] (-0.55,0.06) -- (-0.72,-0.18); \draw[person] (-0.55,0.06) -- (-0.38,-0.18);}
\figperson{0.55}{\draw[person] (0.55,0.06) -- (0.38,-0.18); \draw[person] (0.55,0.06) -- (0.70,-0.02);
  \fill[brown!70!black] (0.69,-0.07) rectangle (0.77,0.05);}
\foreach \x/\n in {-1.65/1, -0.55/2, 0.55/3} {
  \draw[box] (\x-0.30,-0.80) rectangle (\x+0.30,0.48);
  \node[tag, anchor=south west] at (\x-0.30,0.48) {\n};
}
\node[hdr] at (-3.75,-1.30) {Intent only (S-C)};
\node[outb, anchor=north west, text width=2.45cm] at (-3.75,-1.58)
  {aim: [commuting]\hfill{\color{red!70!black}\normalfont\ding{55}}};
\node[hdr] at (-0.72,-1.30) {Actions, then intent (J-FC)};
\node[outb, anchor=north west, text width=4.19cm] at (-0.72,-1.58)
  {1: standing, looking at robot\\2: standing\\3: standing, holding sth\\[1pt]aim: [waiting]\hfill{\color{green!45!black}\normalfont\ding{51}}};
\end{tikzpicture}
\caption{Three people at a bus stop seen by a mobile robot. A model trained only on group intent guesses \emph{commuting}; a model trained to first write each member's actions answers \emph{waiting}.}
\label{fig:example}
\end{figure}

Vision-language models (VLMs) make this question concrete. A single model can be trained to name each member's actions, to name the group's intent, or to write both in one answer, with every other part of the recipe held fixed. This lets us ask three questions that are hard to separate in hand-built hierarchical models. Do the details help the whole? Does the whole help the details? And if one direction helps, what exactly in the details carries the gain?

We study these questions on JRDB-Social \citep{jahangard2024jrdb}, where each annotated group comes with per-member action labels and a group intent label (Figure~\ref{fig:example}). We fine-tune Qwen3.5 4B with three seeds under eight training arms, two single-task references and six arms that combine, mix or condition the two levels, and repeat the core arms on Qwen3.5 9B and Qwen3-VL 4B. A verdict rule written before any test run calls an arm reinforcing only when it beats both single-task references. Around this grid we run controls for prompt wording, learning rate, label budget and the content of the action targets, we extend the comparison to a tool-using agent and to GRPO rewards \citep{shao2024deepseekmath}, and we open the models with five mechanistic analyses.

The answer is one-directional. Writing each member's actions before the intent (\jfc) raises intent macro-F1 from 30.1 to 31.0, and writing the intent first (\jcf) reaches 31.2, while action F1 stays within 0.2 points of the action-only model. Under the selected learning rate, no arm trained with intent improves actions. The gain needs correct actions that match the image: an empty action field lowers intent F1 to 29.8, and actions borrowed from another group with the same intent lower it to 28.1. Forced prefixes show that the intent follows the actions the model writes, occlusion and member masking show that the joint model relies more on people and on each individual member, and grafting places part of the gain in the vision tower. When the same actions are supplied as input during training instead of as targets, both levels drop.

These results matter for anyone who builds social perception on top of a VLM. Asking the model to write out what each person is doing is a cheap way to improve group-level understanding, it costs at most 0.2 action points on Qwen3.5 4B, and it helps most when intent labels are scarce. The same information given as a hint in the prompt does harm. For research on multi-granularity learning, the results show that a gain in one direction can coexist with no gain in the other, so both directions need their own matched reference. Our contributions are:
\begin{itemize}
\setlength\itemsep{0pt}
\item A study of cross-granularity transfer between person actions and group intent in VLMs, with eight training arms, three seeds, a pre-registered verdict rule and controls for prompt, learning rate, label budget and target content.
\item Evidence that the transfer runs from details to the whole only, that it needs correct and visually grounded details, and that it is largest when intent labels are scarce.
\item Analyses that trace the transfer to the written actions, to a stronger reliance on people and on individual members, and partly to the vision tower, together with its limits under conditioning, a second backbone, a tool-using agent and RL rewards.
\end{itemize}

\section{Related Work}
\label{sec:related}

\paragraph{Person and group levels in social scenes.}
Group activity recognition builds group labels from person features, either through temporal hierarchies \citep{ibrahim2016hierarchical}, relation graphs \citep{wu2019learning} and transformers \citep{gavrilyuk2020actor,zhou2022composer}, or without any person labels at all \citep{kim2022detector}. Adding a person action loss raises group accuracy on volleyball and collective activity data \citep{ibrahim2016hierarchical,choi2009they,ehsanpour2020joint}, while on JRDB-PAR coupling the global level to the person level lowers global F1 \citep{han2022panoramic}. JRDB-Social \citep{jahangard2024jrdb} evaluates VLMs zero shot at the person, pair and group levels and finds that accuracy falls from the person to the group level. A recent VLM system injects a coarse description of an interaction as a prior for fine answers \citep{su2026earl}.

\paragraph{Mutual reinforcement across granularities.}
Multi-task learning shares structure across related tasks \citep{caruana1997multitask,ruder2017overview}, and the benefit can be uneven between task pairs \citep{standley2020tasks,fifty2021efficiently}, partly because task gradients conflict \citep{yu2020gradient}. In multimodal models, region-level grounding and image-level answering have been reported to help each other \citep{you2024ferret,chen2023shikra,shao2024visual}, though joint training can also trade one ability for the other \citep{zhang2024llava,fan2026grit}. Writing intermediate steps before an answer helps language and vision-language models reason \citep{wei2022chain,zhang2023multimodal}, training on such steps as targets helps small models \citep{hsieh2023distilling}, and editing the steps tests whether the answer depends on them \citep{lanham2023measuring}. The mutual reinforcement effect (MRE) states that word-level and text-level labels improve each other when learned together, first for Japanese sentence classification and entity recognition \citep{gan2023sentence} and then across languages and extraction tasks \citep{gan2026multilingual}.

\paragraph{Our position.}
Earlier studies either test one direction or compare against single-task models trained with a different recipe. We test both directions between person actions and group intent inside the same VLM, under one shared recipe, three seeds and a verdict rule fixed in advance, and we follow the result through content controls, mechanistic analyses, agents and rewards. This turns a common modelling assumption into a measured, direction-specific effect that practitioners can act on.

\section{Task Formulation: Person Actions, Group Intent and Eight Training Arms}
\label{sec:task}

\subsection{Person Actions as Details and Group Intent as the Whole}
\label{sec:formulation}

A sample $s$ is one keyframe of one social group. The input $X$ holds up to three frames about two seconds apart, with the $n$ members marked by numbered red boxes, and a zoomed crop of the group in the target frame. Each member $m \in \{1,\dots,n\}$ carries a set of actions $A_m \subseteq \mathcal{A}$ drawn from 27 action labels, and the group carries an intent set $Y \subseteq \mathcal{C}$ drawn from eight headline intents (commuting, eating, socializing, navigating, studying or working, discussing, waiting, leisure). Groups with $n = 1$ are single people with an intent of their own.

The model answers in JSON. We write $\tau_F(s)$ for the serialized member list $\{(m, A_m)\}_{m=1}^{n}$ and $\tau_C(s)$ for the serialized intent list $Y$. A prompt $q$ states which fields to produce. Every arm minimizes a token-level negative log-likelihood of the form
\begin{equation}
\mathcal{L}(\theta) = - \mathbb{E}_{s} \sum_{(q,t) \in \mathcal{T}(s)} \log p_\theta\!\left(t \mid X, q\right),
\label{eq:loss}
\end{equation}
where $\mathcal{T}(s)$ is the set of prompt and target pairs that the arm builds from $s$.

\subsection{Eight Training Arms that Combine, Separate or Condition the Two Granularities}
\label{sec:arms}

The arms differ only in $\mathcal{T}(s)$, as summarized in Figure~\ref{fig:overview}a. The two single-task references are
\begin{align}
\mathcal{T}_{\textsc{S-F}}(s) &= \{(q_F, \tau_F)\}, \\
\mathcal{T}_{\textsc{S-C}}(s) &= \{(q_C, \tau_C)\}.
\end{align}
The joint arms put both levels into one answer, in either order:
\begin{align}
\mathcal{T}_{\textsc{J-FC}}(s) &= \{(q_{FC}, \tau_F \oplus \tau_C)\}, \\
\mathcal{T}_{\textsc{J-CF}}(s) &= \{(q_{CF}, \tau_C \oplus \tau_F)\},
\end{align}
where $\oplus$ concatenates the two JSON fields. Under \jfc, the chain rule splits the loss into $\log p(\tau_F \mid X) + \log p(\tau_C \mid X, \tau_F)$, so the intent is learned and decoded after the model has written the actions; \jcf\ reverses this order. \textsc{MIX} keeps the two levels as separate samples, $\mathcal{T}_{\textsc{MIX}}(s) = \mathcal{T}_{\textsc{S-F}}(s) \cup \mathcal{T}_{\textsc{S-C}}(s)$. The conditioned arms place the other level in the prompt during training:
\begin{equation}
\begin{split}
\mathcal{T}_{\textsc{COND}}(s) = \{&(q_F \oplus c(\tau_C), \tau_F),\\
&(q_C \oplus c(\tau_F), \tau_C)\},
\end{split}
\end{equation}
and are tested with the plain prompts $q_F$ and $q_C$. \textsc{COND} uses the gold labels of the same group for $c(\cdot)$, \textsc{COND-N} fills the slot with ``N/A'', and \textsc{COND-S} copies the labels of a donor group from another location with a different intent and, when possible, the same number of members.

\subsection{A Pre-registered Verdict Rule for Mutual Reinforcement}
\label{sec:verdict}

Let $F_a$ be the action score and $C_a$ the intent score of arm $a$ (\S\ref{sec:setup}), averaged over seeds. For every arm other than the references we compute
\begin{equation}
\Delta_F(a) = F_a - F_{\textsc{S-F}}, \qquad \Delta_C(a) = C_a - C_{\textsc{S-C}},
\end{equation}
and assign the verdict
\begin{equation}
v(a) =
\begin{cases}
\textsc{Reinforces} & \Delta_F > 0,\ \Delta_C > 0,\\
\textsc{Harms} & \Delta_F < 0,\ \Delta_C < 0,\\
\textsc{Trades} & \text{otherwise.}
\end{cases}
\label{eq:verdict}
\end{equation}
We fixed this rule, the references, the metrics and the test split before any test evaluation, and we kept a dated record of every later addition.

\section{Experimental Setup}
\label{sec:setup}

\begin{figure*}[t]
\centering
\resizebox{\textwidth}{!}{
\begin{tikzpicture}[
  font=\scriptsize, x=1cm, y=1cm,
  panel/.style={draw=black!25, rounded corners=3pt, fill=black!2},
  ptitle/.style={font=\scriptsize\bfseries, anchor=north west, inner sep=0pt},
  lab/.style={anchor=base west, yshift=-0.07cm, inner sep=0pt, font=\scriptsize},
  sub/.style={anchor=base west, yshift=-0.07cm, inner sep=0pt, font=\scriptsize\itshape},
  img/.style={draw=black!45, fill=black!8, minimum width=0.42cm, minimum height=0.28cm, inner sep=0pt, rounded corners=0.8pt},
  A/.style={fill=blue!55!black!70, text=white, minimum width=0.30cm, minimum height=0.28cm, inner sep=0pt, font=\tiny\bfseries, rounded corners=0.8pt},
  I/.style={fill=orange!85!black, text=white, minimum width=0.30cm, minimum height=0.28cm, inner sep=0pt, font=\tiny\bfseries, rounded corners=0.8pt},
  Ax/.style={A, fill=blue!20, text=blue!50!black},
  Ab/.style={A, fill=none, preaction={fill=white}, pattern=north east lines, pattern color=blue!40, text=blue!50!black, draw=blue!50!black},
  Ae/.style={draw=blue!50!black, dash pattern=on 1.2pt off 0.8pt, minimum width=0.30cm, minimum height=0.28cm, inner sep=0pt, rounded corners=0.8pt},
  arr/.style={-{Stealth[length=2.2pt]}, black!55, line width=0.45pt},
  note/.style={anchor=base west, yshift=-0.05cm, inner sep=0pt, font=\tiny, text=black!70},
  tool/.style={draw=black!40, rounded corners=1pt, inner sep=1.5pt, font=\tiny, anchor=west},
  sep/.style={black!20, line width=0.4pt},
]
\begin{scope}[yshift=2.9cm]
\fill[panel] (0,0) rectangle (7.9,2.7);
\node[ptitle] at (0.15,2.55) {(a) Eight training arms (\S\ref{sec:arms})};
\foreach \y/\name/\txt in {1.95/S-F/{actions only}, 1.57/S-C/{intent only}, 1.19/J-FC/{actions, then intent}, 0.81/J-CF/{intent, then actions}} {
  \node[lab] at (0.15,\y) {\name};
  \node[img] at (1.15,\y) {};
  \draw[arr] (1.40,\y) -- (1.68,\y);
  \node[note] at (2.45,\y) {\txt};
}
\node[A] at (1.88,1.95) {A};
\node[I] at (1.88,1.57) {I};
\node[A] at (1.88,1.19) {A}; \node[I] at (2.22,1.19) {I};
\node[I] at (1.88,0.81) {I}; \node[A] at (2.22,0.81) {A};
\node[note] at (0.15,0.38) {one prompt and one answer per sample};
\draw[sep] (4.32,0.30) -- (4.32,2.15);
\begin{scope}[xshift=4.50cm]
\node[lab] at (0,1.78) {MIX};
\node[img] at (1.05,1.95) {}; \draw[arr] (1.30,1.95) -- (1.58,1.95); \node[A] at (1.78,1.95) {A};
\node[img] at (1.05,1.61) {}; \draw[arr] (1.30,1.61) -- (1.58,1.61); \node[I] at (1.78,1.61) {I};
\node[note] at (2.08,1.78) {two samples};
\node[lab] at (0,1.07) {COND};
\node[img] at (1.05,1.24) {}; \node[Ax, scale=0.8] at (1.05,1.24) {A}; \draw[arr] (1.30,1.24) -- (1.58,1.24); \node[I] at (1.78,1.24) {I};
\node[img] at (1.05,0.90) {}; \node[I, scale=0.8, fill=orange!30, text=orange!60!black] at (1.05,0.90) {I}; \draw[arr] (1.30,0.90) -- (1.58,0.90); \node[A] at (1.78,0.90) {A};
\node[note] at (2.08,1.07) {gold hint};
\node[note] at (0,0.56) {COND-N: ``N/A'' as the hint};
\node[note] at (0,0.34) {COND-S: another group's labels};
\end{scope}
\end{scope}
\begin{scope}[xshift=8.1cm, yshift=2.9cm]
\fill[panel] (0,0) rectangle (7.9,2.7);
\node[ptitle] at (0.15,2.55) {(b) Controls on the joint target (\S\ref{sec:controls})};
\foreach \y/\name in {1.95/J-nFC, 1.57/J-sFC, 1.19/J-aFC} {
  \node[lab] at (0.15,\y) {\name};
  \node[img] at (1.20,\y) {};
  \draw[arr] (1.45,\y) -- (1.73,\y);
}
\node[Ae] at (1.93,1.95) {}; \node[I] at (2.27,1.95) {I}; \node[note] at (2.55,1.95) {empty lists};
\node[Ab] at (1.93,1.57) {A}; \node[I] at (2.27,1.57) {I}; \node[note] at (2.55,1.57) {donor, other intent};
\node[Ax] at (1.93,1.19) {A}; \node[I] at (2.27,1.19) {I}; \node[note] at (2.55,1.19) {donor, same intent};
\node[note] at (0.15,0.68) {donors come from another location;};
\node[note] at (0.15,0.46) {prompt and format stay those of J-FC};
\draw[sep] (4.32,0.30) -- (4.32,2.15);
\begin{scope}[xshift=4.50cm]
\node[lab] at (0,1.95) {S-C-P}; \node[note] at (0.85,1.95) {intent only, J-FC prompt};
\node[lab] at (0,1.57) {lr}; \node[note] at (0.85,1.57) {$10^{-5}$ and $2{\times}10^{-5}$};
\node[lab] at (0,1.19) {Labels}; \node[note] at (0.85,1.19) {intent on 25\%, 50\%, 100\%};
\node[lab] at (0,0.81) {Steps}; \node[note] at (0.85,0.81) {single task, 6 epochs};
\end{scope}
\end{scope}
\fill[panel] (0,0) rectangle (7.9,2.7);
\node[ptitle] at (0.15,2.55) {(c) Backbones, agents and rewards (\S\ref{sec:beyond})};
\node[sub] at (0.15,1.95) {Backbones};
\node[note] at (0.15,1.57) {Qwen3.5 4B, 3 seeds};
\node[note] at (0.15,1.32) {Qwen3.5 9B, 1 seed};
\node[note] at (0.15,1.07) {Qwen3-VL 4B, 3 seeds};
\draw[sep] (2.32,0.30) -- (2.32,2.15);
\node[sub] at (2.50,1.95) {Tool-using agent};
\node[tool] (f) at (2.50,1.50) {frames};
\node[tool] (z) at (3.45,1.50) {zoom $k$};
\draw[arr, shorten >=0.6pt] (f) -- (z);
\draw[arr] (z.east) -- (4.62,1.50);
\node[A] at (4.82,1.50) {A}; \node[I] at (5.16,1.50) {I};
\node[note] at (2.50,1.02) {AG-F, AG-C, AG-FC, AG-CF};
\node[note] at (2.50,0.80) {share tools and images and};
\node[note] at (2.50,0.58) {differ only in their notes};
\draw[sep] (5.58,0.30) -- (5.58,2.15);
\node[sub] at (5.76,1.95) {GRPO from J-FC};
\node[note] at (5.76,1.57) {$r = R_C$};
\node[note] at (5.76,1.32) {$r = R_F$};
\node[note] at (5.76,1.07) {$r = R_C + \tfrac{1}{2} R_F$};
\node[note] at (5.76,0.80) {gated variant};
\node[note] at (5.76,0.58) {agent version};
\begin{scope}[xshift=8.1cm]
\fill[panel] (0,0) rectangle (7.9,2.7);
\node[ptitle] at (0.15,2.55) {(d) Analyses (\S\ref{sec:analysis})};
\node[sub] at (0.15,1.95) {Forced prefix};
\node[A] at (0.35,1.50) {A}; \node[note] at (0.62,1.50) {forced: correct, wrong, empty};
\draw[arr] (0.35,1.33) -- (0.35,1.14);
\node[I] at (0.35,0.96) {I}; \node[note] at (0.62,0.96) {decoded and scored};
\node[note] at (0.15,0.52) {and the reverse order in J-CF};
\draw[sep] (3.35,0.30) -- (3.35,2.15);
\begin{scope}[xshift=3.55cm]
\node[lab] at (0,1.95) {Occlusion}; \node[note] at (1.95,1.95) {people only, place only};
\node[lab] at (0,1.57) {Member masking}; \node[note] at (1.95,1.57) {hide one member};
\node[lab] at (0,1.19) {Linear probes}; \node[note] at (1.95,1.19) {16 targets, all layers};
\node[lab] at (0,0.81) {Grafting}; \node[note] at (1.95,0.81) {swap weight groups};
\node[note] at (0,0.40) {\textcolor{blue!55!black}{\textbf{A}} person actions\qquad \textcolor{orange!85!black}{\textbf{I}} group intent};
\end{scope}
\end{scope}
\end{tikzpicture}}
\caption{Overview of the study. (a) The training arms differ only in which levels appear as targets and in which order. (b) Controls on the joint target, the prompt, the learning rate, the intent labels and the training steps. (c) Other backbones, a tool-using agent and GRPO rewards. (d) Analyses inside the trained models.}
\label{fig:overview}
\end{figure*}
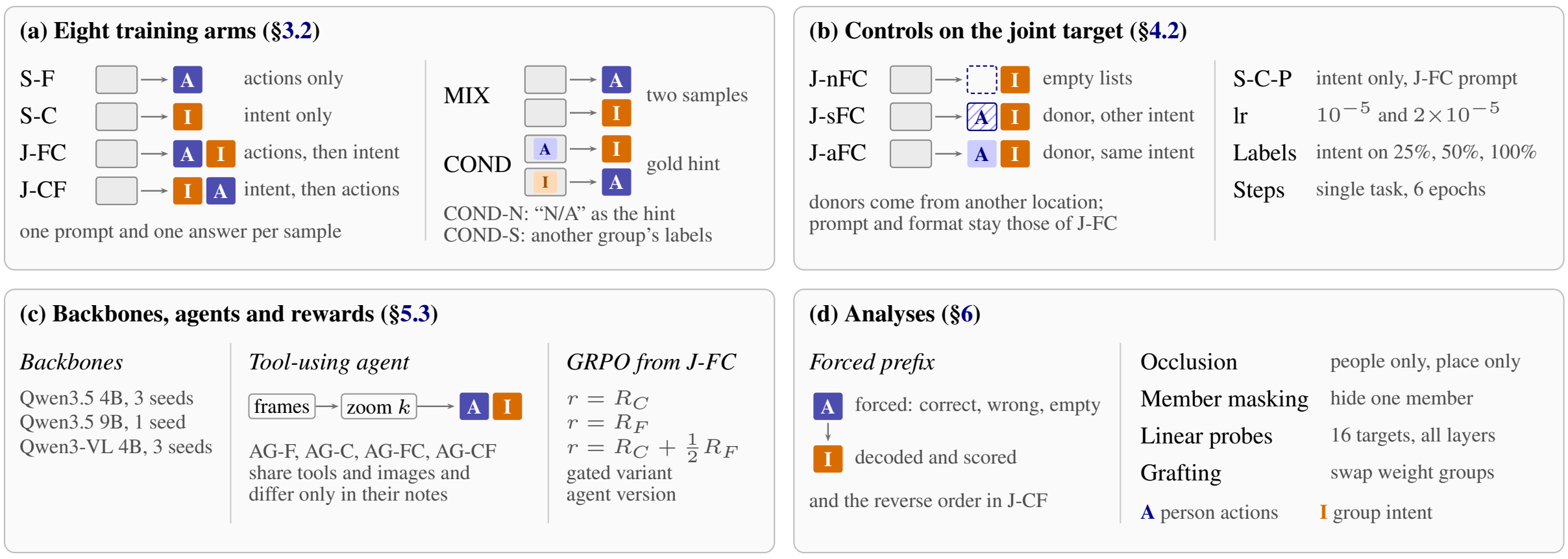

\subsection{JRDB-Social, Backbones and Evaluation Metrics}
\label{sec:data}

\paragraph{Data.}
We use the official JRDB-Social split \citep{jahangard2024jrdb} of 20 training, 7 development and 27 test sequences recorded by a mobile robot in 19 test locations. Each group contributes up to six keyframes for training and up to five for evaluation, which gives 4{,}540 training, 1{,}397 development and 5{,}427 test samples. The test set contains 1{,}136 groups, 250 with several members and 886 single people. After selecting the learning rate on the development set, every final model is trained on training and development data together.

\paragraph{Backbones and training.}
The main backbone is Qwen3.5 4B \citep{team2026qwen3}, trained with seeds 42, 1 and 2. We repeat the core arms on Qwen3.5 9B (seed 42) and on Qwen3-VL 4B Instruct \citep{bai2025qwen3} (three seeds). All arms share one recipe: full fine-tuning for three epochs with learning rate $10^{-5}$, chosen on the development set by the summed scores of \sF\ and \sC, and the same batch size, schedule and image resolution. Decoding is greedy, and every test answer of every supervised arm parses as valid JSON.

\paragraph{Metrics.}
The action score $F$ is the macro-F1 over the set $\mathcal{A}_{12}$ of the 12 actions with at least 20 test instances,
\begin{equation}
F = \frac{1}{|\mathcal{A}_{12}|} \sum_{c \in \mathcal{A}_{12}} \frac{2\,\mathrm{TP}_c}{2\,\mathrm{TP}_c + \mathrm{FP}_c + \mathrm{FN}_c},
\end{equation}
where the counts run over all (member, action) pairs of the test samples and actions that the annotators marked as hard are ignored on both sides. The intent score $C$ is a group-level macro-F1 over the eight headline intents. A group's predicted intent set is the majority vote over its keyframes,
\begin{equation}
\hat{Y}_g = \Bigl\{ c : \textstyle\sum_{k=1}^{K_g} \mathbf{1}[c \in \hat{Y}_{g,k}] \ge K_g / 2 \Bigr\},
\end{equation}
where $K_g$ is the number of keyframes of group $g$. We also report $C$ on multi-member and single-person groups, on five merged intent families, and on 336 groups whose intent a human or a strong VLM judged recoverable from the frames, a subset whose definition was fixed before any Qwen model was evaluated (Appendix~\ref{app:data}).

\subsection{Controls for Prompt Wording, Learning Rate, Label Budget and Detail Content}
\label{sec:controls}

Five controls, shown in Figure~\ref{fig:overview}b, test what the joint target adds. \emph{Content controls} keep the \jfc\ prompt and output format and change only the action part of the target: \textsc{J-nFC} writes an empty action list for every member, \textsc{J-sFC} copies the actions of a donor group with another intent, and \textsc{J-aFC} copies the actions of a donor group from another location with the same intent. These arms separate the effect of correct actions from the effect of the output format and from the effect of intent-consistent action text. \emph{Prompt matching} trains \textsc{S-C-P}, an intent-only model that receives the \jfc\ instruction. \emph{Learning rate} repeats \sF, \sC, \jfc\ and \textsc{COND} at $2 \times 10^{-5}$. \emph{Label budget} keeps intent labels for 25\% or 50\% of the training groups, with two independent draws per budget, while all action labels stay. \emph{Training steps} trains \sF\ and \sC\ for six epochs to match the number of updates of \textsc{MIX} and \textsc{COND}.

\section{Results: Person Actions Lift Group Intent, but Not the Reverse}
\label{sec:results}

\begin{table*}[t]
\centering
\small
\setlength{\tabcolsep}{5pt}
\begin{tabular}{l rr rr rrrr l}
\toprule
& \multicolumn{2}{c}{Person actions} & \multicolumn{6}{c}{Group intent} & \\
\cmidrule(lr){2-3}\cmidrule(lr){4-9}
Arm & F1 & $\Delta_F$ & F1 & $\Delta_C$ & Multi & Single & 5-class & Visible & Verdict \\
\midrule
Zero shot & 34.8 &  & 26.5 &  & 23.6 & 25.9 & 37.5 & 29.1 & zero shot \\
S-F & 59.8 &  & -- &  & -- & -- & -- & -- & reference \\
S-C & -- &  & 30.1 &  & 31.6 & 25.0 & 48.7 & 31.8 & reference \\
\midrule
MIX & 57.6 & $-$2.2 & 30.6 & +0.5 & 32.0 & 25.5 & 48.9 & 33.3 & \textsc{Trades} \\
J-FC & 59.7 & $-$0.1 & 31.0 & +0.9 & 31.6 & 25.9 & 48.3 & 33.4 & \textsc{Trades} \\
J-CF & 59.6 & $-$0.2 & 31.2 & +1.1 & 31.4 & 26.1 & 48.8 & 33.8 & \textsc{Trades} \\
COND & 57.0 & $-$2.8 & 27.8 & $-$2.3 & 28.4 & 24.3 & 45.5 & 29.0 & \textsc{Harms} \\
COND-N & 58.6 & $-$1.2 & 30.1 & +0.0 & 31.7 & 24.5 & 48.6 & 32.4 & \textsc{Trades} \\
COND-S & 58.3 & $-$1.5 & 29.5 & $-$0.6 & 30.1 & 25.1 & 47.4 & 32.2 & \textsc{Harms} \\
\bottomrule
\end{tabular}
\caption{Main results on the JRDB-Social test set with Qwen3.5-4B (macro-F1 $\times 100$, mean of three seeds). $\Delta_F$ and $\Delta_C$ compare each arm with S-F and S-C. Multi and Single split the intent score by group size, 5-class merges the intent labels into five families, and Visible restricts it to the 336 groups whose intent can be judged from the frames (\S\ref{sec:setup}).}
\label{tab:main}
\end{table*}

\subsection{Joint Training Lifts Group Intent while Person Actions Stay Level}
\label{sec:main}

Table~\ref{tab:main} gives the main grid. Both joint arms raise intent F1 over \sC, by 0.9 points for \jfc\ and 1.1 points for \jcf, and keep action F1 within 0.2 points of \sF. \textsc{MIX} also gains on intent (+0.5) but loses 2.2 points on actions, so presenting the two levels as separate samples costs the details without adding much to the whole. Under the selected recipe no arm reaches \textsc{Reinforces}: every arm that sees intent labels scores below \sF\ on actions. The transfer therefore runs in one direction, from the details to the whole.

On the 336 groups whose intent is judged visible in the frames, \jfc, \jcf\ and \textsc{MIX} add 1.7, 2.0 and 1.5 points over \sC. On the full test set the gain comes from single people (+0.9 and +1.1 for \jfc\ and \jcf), where the intent of the ``group'' is the intent of one person, while multi-member groups stay level; on single people \jfc\ pays for it with 1.5 action points (Appendix~\ref{app:perseed}). On five merged intent families the joint arms stay within 0.4 points of \sC, so the gain sits in the finer distinctions within a family, which \S\ref{sec:content} examines class by class. Qwen3.5 9B shows the same pattern with a single seed: \jfc\ gains 1.5 intent points and loses 1.5 action points.

Three controls test whether something other than the action targets explains the gain. \emph{Prompt matching}: \textsc{S-C-P}, which reads the \jfc\ instruction but answers only the intent, scores 29.6, so the instruction alone does not help and \jfc\ is 1.4 points above it; its counterpart \textsc{S-F-P} scores 60.4 on actions, the same as \sF\ (seed 42). \emph{Training steps}: six epochs of single-task training (seed 42) lower both references (\sF\ 59.3, \sC\ 28.7), so the extra updates of \textsc{MIX} and \textsc{COND} do not explain their scores; against these step-matched references \textsc{MIX} still trades ($-0.9$ / $+2.2$) and \textsc{COND} still harms ($-1.6$ / $-0.7$). \emph{Learning rate}: at $2 \times 10^{-5}$ (seed 42) the single-task references fall to 57.8 and 27.9 while \jfc\ falls only to 58.5 on actions and holds 31.2 on intent, which makes \jfc\ better than both references and turns its verdict into \textsc{Reinforces} for that learning rate. \textsc{COND} also moves from \textsc{Harms} to \textsc{Trades} ($-0.7$ / $+0.4$): at this learning rate the references degrade more than the other arms.

\begin{figure}[t]
\centering
\includegraphics[width=\columnwidth]{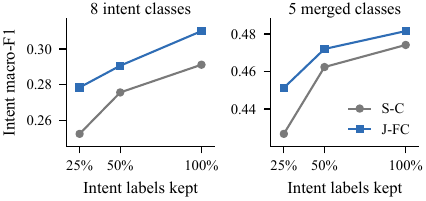}
\caption{Intent macro-F1 when intent labels are kept for 25\%, 50\% or all training groups (seed 42; two draws per budget, averaged). Action labels are always complete.}
\label{fig:budget}
\end{figure}

\emph{Label budget.} When intent labels are kept for only 25\% of the training groups (seed 42), \sC\ drops to 25.3 on average over two draws while \jfc, which still sees every action label, holds 27.9 (Figure~\ref{fig:budget}). The gap is 2.6 points at 25\%, 1.5 at 50\% and 1.9 at full labels, and on the five merged families it grows from 0.7 to 1.0 to 2.5 as labels become scarce. \jfc\ never falls below \sC\ in any of the five comparisons.

\subsection{Only Correct and Visually Grounded Details Carry the Gain}
\label{sec:content}

\begin{figure}[t]
\centering
\includegraphics[width=\columnwidth]{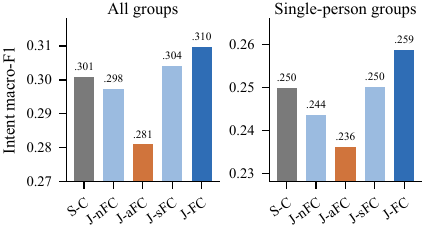}
\caption{Intent macro-F1 (three seeds) when the action part of the joint target is empty (J-nFC), taken from another group with the same intent (J-aFC), taken from another group with a different intent (J-sFC), or correct (J-FC).}
\label{fig:content}
\end{figure}

The content controls in Figure~\ref{fig:content} keep the \jfc\ instruction and output format and change only what the action field contains. Writing an empty action list for every member gives 29.8, slightly below \sC. Copying the actions of another group with the same intent gives 28.1, the lowest of the content controls. Only the correct actions of the group in the image reach 31.0. The gain of \jfc\ thus decomposes as
\begin{equation}
\begin{split}
\underbrace{C_{\textsc{J-FC}} - C_{\textsc{S-C}}}_{+0.9}
&= \underbrace{C_{\textsc{J-FC}} - C_{\textsc{J-nFC}}}_{+1.2\ \text{content}} \\
&\quad + \underbrace{C_{\textsc{J-nFC}} - C_{\textsc{S-C}}}_{-0.3\ \text{format}},
\end{split}
\end{equation}
so the output format contributes nothing and the action content carries the whole gain. The comparison with \textsc{J-aFC} matters most: those targets are consistent with the intent label and differ from the correct ones only in not describing the people in the image, and they cost 2.9 points against \jfc. What helps the whole is a description of what these particular people are doing, not a list of actions that usually go with the intent. Actions from a group with a different intent (\textsc{J-sFC}) end in between at 30.4. In every subset, correct actions score above empty actions and above same-intent donor actions (Appendix Table~\ref{tab:content}).

\begin{figure*}[t]
\centering
\includegraphics[width=\textwidth]{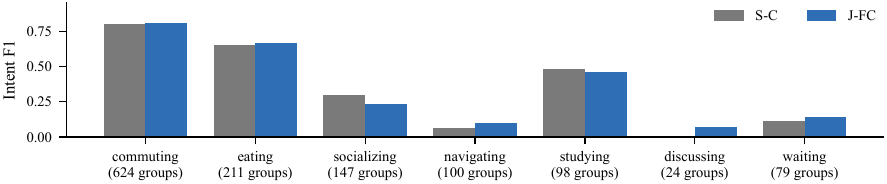}
\caption{Per-class intent F1 of \sC\ and \jfc, averaged over three seeds. Leisure is 0 for both and is not shown.}
\label{fig:perclass}
\end{figure*}

\emph{Which intents gain.} The gain falls on a few small intents (Figure~\ref{fig:perclass}). Per class, \jfc\ raises navigating from 6.1 to 9.9, waiting from 11.3 to 14.3 and discussing from 0.0 to 6.8, while commuting moves from 80.4 to 80.6 and eating rises by 1.5 points. Of the macro average, discussing adds 0.85 points, navigating 0.48 and waiting 0.37, while socializing takes away 0.78. \sC\ predicts on average 6.3 distinct intents over the test set against 7.0 for \jfc, so single-task training collapses part of the rare intents and action supervision keeps them alive. Socializing loses recall (Appendix~\ref{app:perclass}).

\subsection{Details as Input, a Second Backbone, Tool-Using Agents and RL Rewards}
\label{sec:beyond}

\begin{table}[t]
\centering
\small
\setlength{\tabcolsep}{4pt}
\begin{tabular}{ll rr l}
\toprule
Setting & Arm & $\Delta_F$ & $\Delta_C$ & Verdict \\
\midrule
Qwen3-VL 4B & MIX & $-$2.0 & +0.5 & \textsc{Trades} \\
 & J-FC & $-$0.8 & $-$1.2 & \textsc{Harms} \\
 & J-CF & $-$1.7 & +0.0 & \textsc{Trades} \\
 & COND & $-$4.2 & $-$1.1 & \textsc{Harms} \\
\midrule
Qwen3.5 9B$^\dagger$ & MIX & $-$3.5 & +1.0 & \textsc{Trades} \\
 & J-FC & $-$1.5 & +1.5 & \textsc{Trades} \\
 & COND & $-$5.4 & $-$2.7 & \textsc{Harms} \\
\midrule
Agent (4B) & AG-FC & $-$0.3 & $-$0.5 & \textsc{Harms} \\
 & AG-CF & $-$1.6 & $-$2.1 & \textsc{Harms} \\
\midrule
GRPO (4B)$^\ddagger$ & $R_C$ & $-$0.1 & +0.8 & -- \\
 & $R_F$ & $-$1.1 & +0.0 & -- \\
 & $R_C{+}\frac{1}{2}R_F$ & $-$1.0 & +0.1 & -- \\
\bottomrule
\end{tabular}
\caption{The transfer outside the main setting ($\times 100$). Deltas use the single-task arms of the same setting as references; the agent rows compare with AG-F and AG-C. $^\dagger$One seed. $^\ddagger$Two seeds, deltas relative to the J-FC starting point.}
\label{tab:boundaries}
\end{table}

Table~\ref{tab:boundaries} moves the comparison away from the main setting. \emph{Details as input.} The \textsc{COND} arm gives the model the gold actions in its training prompt and tests it without them, much like learning with privileged information \citep{vapnik2009new}. It lowers both levels on all three backbones, by 2.8 and 2.3 points on Qwen3.5 4B, and the harm follows the quality of the hint: gold labels cost 2.3 intent points, a donor group's labels 0.6 and ``N/A'' nothing. A model that learns to read the answer from its prompt stops looking for it in the image, a shortcut in the sense of \citet{geirhos2020shortcut}.

\emph{A second backbone.} On Qwen3-VL 4B, \textsc{MIX} and \jcf\ still raise intent over \sC\ (by 0.5 and 0.05 points) while losing actions, so the direction of the transfer holds, but its size shrinks and \jfc\ falls on both levels.

\emph{Tool-using agents.} We give the 4B model an offline harness with zoom and frame tools in the style of ReAct \citep{yao2022react} and train four agent arms on scripted trajectories that differ only in the notes they write (Appendix~\ref{app:agent}). The tools alone already lift intent: the intent-only agent \textsc{AG-C} reaches 30.9 against 30.1 for \sC. The joint agent arms then lose to the single-task agents on both levels over all groups, while on single people \textsc{AG-FC} still gains 1.0 intent points. A likely reason is that once the agent zooms into the members, the image evidence that \jfc\ writes down is already in its context.

\emph{RL rewards.} GRPO started from \jfc\ raises intent by 0.8 points with an intent reward and by 0.1 when half an action reward is added, and an action reward alone lowers actions by 1.1 points. When the supervised model is trained without the development split and GRPO runs on development prompts (seed 42), every reward with an intent term lowers intent, by 1.8 to 3.3 points, although the rewards rise on the RL prompts. The transfer comes from the supervised targets, and an action reward adds nothing to intent (Appendix~\ref{app:grpo}), in line with \citet{chen2026does}.

\section{Analysis: How Group Intent Draws on Person Actions}
\label{sec:analysis}

\subsection{Group Intent Follows the Actions the Model Writes}
\label{sec:prefix}

\begin{figure}[t]
\centering
\includegraphics[width=\columnwidth]{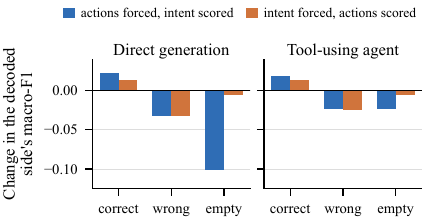}
\caption{Forced prefixes. Blue: the model's action field is replaced by the correct actions, another group's actions or empty lists, and the intent is decoded. Orange: the intent field is replaced and the actions are decoded. Direct generation uses \jfc\ and \jcf\ (three seeds); the agent uses \textsc{AG-FC} and \textsc{AG-CF} with all earlier notes rewritten (seed 42).}
\label{fig:prefix}
\end{figure}

A joint model writes one level before the other, which lets us intervene on the first field and read the second, a mediation test \citep{vig2020investigating} of the kind used to check whether written reasoning drives an answer \citep{lanham2023measuring,turpin2023language}. For a prefix variant $v \in \{\text{correct}, \text{wrong}, \text{empty}\}$ we force the first field, let the model complete the answer, and measure
\begin{equation}
\delta_v = S\bigl(\text{answer} \mid v\bigr) - S\bigl(\text{own answer}\bigr),
\end{equation}
where $S$ is the intent score when actions are forced in \jfc\ and the action score when the intent is forced in \jcf. Wrong prefixes come from a donor group with a different intent, mapped onto the member ids. Forcing the model's own answer as the prefix changes the scores by at most 0.2 points, so the procedure itself is neutral.

Figure~\ref{fig:prefix} shows a clear asymmetry. With actions forced, correct actions raise intent F1 by 2.3 points, wrong actions lower it by 3.2 and empty actions lower it by 10.1; two seeds lose about 4 points without actions, and the third loses 22.6 because it then answers commuting for almost every sample. With the intent forced, the correct intent raises action F1 by 1.3 and a wrong intent lowers it by 3.3, but an empty intent changes it by only 0.5. The intent depends on the written actions, while the actions do not need the intent and only suffer when it is wrong.

The two directions also differ in how far a perfect prefix can carry the other level (seed 42). With correct actions written, \jfc\ reaches 33.9 intent F1, 2.9 points above its own answers and close to the 35.1 that a linear classifier reaches from the gold actions and the group size (Appendix~\ref{app:data}). With the correct intent written, \jcf\ reaches 60.4 action F1, the same as \sF: even a perfect whole adds nothing to the details. The agent behaves the same way once all earlier notes are rewritten as well: empty actions cost the intent 2.3 points and an empty intent costs the actions 0.5.

The decoding order is only part of the story. \jcf\ writes the intent first, cannot read its own actions when it does so, and still gains 1.1 points, so training on the actions also changes how the model sees the scene. The next two analyses look at that change.

\subsection{Joint Models Rely More on People and on Each Member}
\label{sec:people}

\begin{table}[t]
\centering
\small
\setlength{\tabcolsep}{4pt}
\begin{tabular}{l rr}
\toprule
Measure & S-C & J-FC \\
\midrule
Person share $\rho$ of the occlusion loss & 0.40 & 0.46 \\
Intent change, one member hidden & 0.163 & 0.194 \\
Marked member hit rate & 0.545 & 0.555 \\
\quad chance level & 0.471 & 0.477 \\
Action probe gain (S-F: 2.7) & 0.5 & 2.6 \\
Intent probe gain (S-F: 1.1) & 2.0 & 1.8 \\
\bottomrule
\end{tabular}
\caption{How the two models use people and members. $\rho$ is a three-seed mean; the masking and probe rows use seed 42. The hit rate is the share of the human-marked member among the members whose removal deletes a correct intent. Probe gains are balanced accuracy $\times 100$ over the zero-shot model at the best layer, averaged over eight targets.}
\label{tab:mechanism}
\end{table}

\emph{People or place.} We blank either the members (keeping the background) or everything around them (keeping the members) at test time and measure the intent loss in each case. The person share of the loss is
\begin{equation}
\rho = \frac{C - C_{\text{no people}}}{(C - C_{\text{no people}}) + (C - C_{\text{no place}})}.
\end{equation}
Averaged over three seeds, $\rho$ rises from 0.40 for \sC\ to 0.46 for \jfc, 0.50 for \textsc{MIX}, 0.51 for \jcf\ and 0.59 for \textsc{COND} (Table~\ref{tab:mechanism}). The zero-shot model sits at 0.42, so intent supervision alone leaves the balance between people and place where it was, and every arm that learns from actions moves it toward the people. A blurred fill gives the same order (Appendix Table~\ref{tab:i1}).

\emph{Which members.} We then hide one member at a time in multi-member samples (1{,}926 member and sample pairs, seed 42) and record how much the predicted intent set changes,
\begin{equation}
\Delta_k = 1 - \mathrm{F1}\bigl(\hat{Y}, \hat{Y}^{(-k)}\bigr),
\end{equation}
where $\hat{Y}^{(-k)}$ is the prediction with member $k$ hidden. Hiding a single member changes the intent of \jfc\ by 0.194 on average and that of \sC\ by 0.163. A human reviewer marked, for each correct intent, the member who shows it most clearly and rated the others as strong or weak evidence. Among the members whose removal deletes a correct intent, the marked member is found 55.5\% of the time for \jfc\ and 54.5\% for \sC, against chance levels of 47.7\% and 47.1\%, and hiding strong members moves \jfc\ more than hiding weak ones (0.233 against 0.197). Joint training makes the intent depend more on each member while keeping the same preference for the members a person would point to.

\subsection{The Transfer Appears in Hidden States and in the Vision Tower}
\label{sec:inside}

\emph{Linear probes.} We feed 3{,}000 test samples with the same \jfc\ prompt to every model, take the last prompt token at each of the 33 layers, and train logistic probes \citep{alain2016understanding} for eight intent targets and eight action targets with folds split by sequence. For arm $a$ and target set $\mathcal{G}$ we report the gain over the zero-shot model $R_0$ at the best layer,
\begin{equation}
g_a(\mathcal{G}) = \frac{1}{|\mathcal{G}|} \sum_{t \in \mathcal{G}} \bigl( \mathrm{BA}^{\star}_{a}(t) - \mathrm{BA}^{\star}_{R_0}(t) \bigr),
\end{equation}
where $\mathrm{BA}^{\star}(t)$ is the balanced accuracy of target $t$ at the best of the 33 layers. On average over the eight targets, all models are far above a location-only probe and above label permutations \citep{hewitt2019designing} (Appendix~\ref{app:mechanism}). Action supervision adds 2.7 points of action information, while intent supervision adds 0.5 (Table~\ref{tab:mechanism}). Both kinds of supervision add intent information, by 1.1 points for \sF\ and 2.0 for \sC, and \sF\ reaches 3.4 points when probed with its own prompt. Learning the details makes the hidden state richer in both levels; learning the whole leaves the details where the zero-shot model had them.

\emph{Component grafting.} Fine-tuned skills can often be localized in a small part of the weights \citep{panigrahi2023task,ilharco2022editing}. We copy one group of weights from \sC\ into \jfc, or the reverse, and evaluate the result. The groups are the 24 linear-attention layers, the 8 softmax-attention layers, the MLPs, the vision tower and the vision merger. Each graft has three random controls that perturb the same tensors by the same norm in a random direction. Moving the vision tower of \sC\ into \jfc\ lowers intent F1 by 1.05 points, while all three random controls raise it slightly; moving the vision tower of \jfc\ into \sC\ raises intent F1 by 1.13 points, more than seven times the largest random effect (Appendix Figure~\ref{fig:grafting}). Measured as the share of the gap between the two models that a graft moves,
\begin{equation}
m = \frac{C_{\text{graft}} - C_{\text{base}}}{C_{\text{donor}} - C_{\text{base}}},
\end{equation}
the vision tower moves 56\% of the gap in one direction and 60\% in the other. No other component moves intent F1 toward the donor in both directions, and waiting is the only class that follows the vision tower both ways (from 9.7 to 14.3 and from 15.5 to 10.6). Part of the advantage of \jfc\ is stored in how it encodes the image, which fits the content controls: the actions help only when they describe what the image shows.

\section{Conclusion}
\label{sec:conclusion}

Trained on JRDB-Social under one recipe and a verdict rule fixed in advance, Qwen3.5 4B models that write each member's actions together with the group intent gain on intent and keep their action score: the details help the whole, while the whole does not help the details. The gain needs correct actions of the people in the image, the intent reads the actions the model writes, and part of the advantage sits in the vision tower; the same actions given as a training hint lower both levels. For social perception with VLMs this suggests a simple recipe: ask the model what each person is doing before asking what the group is doing, and keep person labels as targets rather than inputs.

\section*{Limitations}

Our study covers one dataset, JRDB-Social, recorded on one university campus, and two model families of 4B and 9B parameters. The 9B model was trained with one seed, and the grafting, probing and member masking analyses use seed 42 only. Group intent in JRDB-Social is hard to judge from a few frames: human agreement with the gold labels reaches a kappa of 0.33, so absolute intent scores are low for every model and the gains we report are about one point on the full test set. Split by test location, the gain of \jfc\ sits in the 58 groups recorded at a training location on another day; it is level with \sC\ on sequences from the same day and location as training data, and 1.0 point below \sC\ on the 427 groups from unseen locations (Appendix~\ref{app:perclass}). The agent and GRPO experiments use one harness and one reward design each.

\bibliography{custom}

\appendix
\setcounter{dbltopnumber}{3}
\renewcommand{\dbltopfraction}{0.9}
\renewcommand{\topfraction}{0.9}
\renewcommand{\textfraction}{0.07}
\section{Data, Labels and Evaluation}
\label{app:data}

\paragraph{Rendering.}
JRDB provides a stitched $3760 \times 480$ panorama per timestep. For each group we cut a $1{,}128 \times 480$ window centred on its members; a group that spans more than 1{,}028 pixels is cut with a 100 pixel margin on each side and scaled to that size. We draw a numbered red box around every member and add a zoomed crop of the group. Multi-frame samples hold the frames about two seconds before and after the target frame, so the model sees up to four images. Member ids follow the left-to-right order in the target frame, and keyframes in which a group has more than eight members are skipped.

\paragraph{Labels.}
JRDB-Social gives every member action a difficulty from 1 to 3. We count actions of difficulty 1 or 2 as positives and treat difficulty 3 as \emph{don't care}: such an action is removed from both the gold and the predicted set of that member. The 12 actions with at least 20 test positives are walking, standing, sitting, holding something, listening to someone, talking to someone, looking at the robot, looking into something, cycling, going upstairs, bending and interacting with a door. The model may answer with any of the 27 actions and nine intents; the ninth intent, attending class, has two test groups and is left out of the macro average. The five merged families are movement (commuting, navigating, leisure), social (socializing, discussing), task (studying or working, attending class), eating and waiting. Table~\ref{tab:classes} lists how many test groups carry each intent.

\begin{table*}[t]
\centering
\small
\setlength{\tabcolsep}{4.5pt}
\begin{tabular}{l rr rr rrr}
\toprule
& \multicolumn{2}{c}{Test groups} & \multicolumn{2}{c}{Oracle F1} & \multicolumn{3}{c}{Cohen's $\kappa$ on blind items} \\
\cmidrule(lr){2-3} \cmidrule(lr){4-5} \cmidrule(lr){6-8}
Intent & all & multi & actions & actions + location & human vs.\ gold & human vs.\ Opus & Opus vs.\ gold \\
\midrule
commuting & 624 & 90 & 83.6 & 79.4 & 0.61 & 0.71 & 0.45 \\
eating & 211 & 76 & 47.0 & 53.6 & 0.66 & 0.69 & 0.63 \\
socializing & 147 & 95 & 47.9 & 47.9 & 0.46 & 0.50 & 0.36 \\
navigating & 100 & 16 & 16.6 & 27.2 & 0.00 & 0.00 & $-0.04$ \\
studying/working & 98 & 36 & 38.4 & 43.6 & 0.69 & 0.61 & 0.53 \\
discussing & 24 & 21 & 13.8 & 12.1 & $-0.03$ & 0.79 & 0.22 \\
waiting & 79 & 16 & 25.4 & 31.9 & 0.11 & 0.46 & 0.16 \\
leisure & 33 & 10 & 8.3 & 12.3 & 0.11 & 0.56 & 0.19 \\
\midrule
macro & 1{,}136 & 250 & 35.1 & 38.5 & 0.33 & 0.54 & 0.31 \\
\bottomrule
\end{tabular}
\caption{The eight headline intents. Oracle F1: a logistic regression on the gold actions and the group size, alone or with the location. $\kappa$: 89 blind items (59 multi-member, 30 single); the human never used the label navigating on them, so its $\kappa$ of 0 is degenerate.}
\label{tab:classes}
\end{table*}

\paragraph{Human review and the visual subset.}
One annotator reviewed all renderable multi-member test groups (251 by the group size in the labels) and 200 single-person test groups sampled by class, 451 in total. For a blind subset the annotator first named the intent without seeing the gold label or any model output; 89 of these items have a blind answer. On them, agreement between the annotator and the gold label is $\kappa = 0.33$ \citep{cohen1960coefficient}, and agreement between the annotator and Claude Opus 5.5 given the same frames is $\kappa = 0.54$ (Table~\ref{tab:classes}). The three intents where \jfc\ gains most (navigating, discussing, waiting) have a human agreement with the gold label of at most 0.11. Where the blind answer or the Opus answer differed from the gold label, the annotator also judged the gold intent as correct, plausible alongside another answer, wrong, or not judgeable. The visual subset contains the 336 groups (205 multi-member, 131 single) where the annotator found the gold intent correct or plausible, or where Opus named a gold intent and the annotator did not mark the gold label as wrong or not judgeable. Its definition was fixed before any Qwen model was evaluated. The subset is 61\% multi-member, against 22\% on the full test set.

\paragraph{Label priors and oracles.}
Predicting the most frequent intent for every group gives 8.9 intent F1 (6.6 on multi-member groups, 9.4 on single people, 16.0 on the five families), and predicting the most frequent intent of each training location gives 15.9 (16.8, 15.1 and 26.8); ten of the 19 test locations do not occur in training. The same two priors reach 4.8 and 8.2 action F1. The zero-shot Qwen3.5 4B reaches 26.5 intent F1. To see how much of the intent is readable from the actions, we fit one class-balanced logistic regression per intent \citep{silla2011survey} on the training and development groups, using the share of members with each gold action and the group size. This oracle reaches 35.1 (35.1 on multi-member groups, 27.7 on single people), and adding the location raises it to 38.5 (43.2 and 29.3). Only 17\% of the test groups get their full intent set right from gold actions, and 36\% with the location added, so actions constrain the intent without fixing it.

\section{Prompts and Training Recipe}
\label{app:prompts}

All arms share one system prompt:

\begin{footnotesize}
\begin{verbatim}
You analyse social groups in frames from a
mobile robot's 360-degree camera. The people
marked with numbered red boxes form one social
group; unmarked people are not in the group.
Member actions (a member can have several):
<27 actions>.
Group aims (a group can have several):
<9 intents>.
Answer with JSON only.
\end{verbatim}
\end{footnotesize}

\noindent The user turn starts with a sentence on the images and the number of members, followed by one of four requests:

\begin{footnotesize}
\begin{verbatim}
F:    For every member, list their actions in
      the target frame. Output:
      {"members": [{"id": 1, "actions":
      ["..."]}, ...]}
C:    What is the group's aim? Output:
      {"aim": ["..."]}
J-FC: For every member, list their actions in
      the target frame, then give the group's
      aim. Output: {"members": [...],
      "aim": ["..."]}
J-CF: Give the group's aim, then for every
      member list their actions in the target
      frame. Output: {"aim": ["..."],
      "members": [...]}
\end{verbatim}
\end{footnotesize}

\noindent The conditioned arms append one sentence during training, ``The group's aim is: \ldots'' to the F request and ``Member actions in the target frame: 1: \ldots; 2: \ldots'' to the C request. \textsc{S-C-P} and \textsc{S-F-P} use the J-FC request with an intent-only or action-only target. For the content controls, the donor of \textsc{J-sFC} and \textsc{COND-S} is a group from another location with a different intent and, when possible, the same number of members; the donor of \textsc{J-aFC} shares the intent instead. Donor members are mapped onto the ids $1,\dots,n$ cyclically, and donors are drawn deterministically. In 14.7\% of the \textsc{J-sFC} training samples the donor actions happen to equal the group's own gold actions.

All supervised runs follow visual instruction tuning \citep{liu2023visual} with full fine-tuning in bf16 and the loss on the answer tokens only, three epochs, a cosine schedule with 3\% warmup and an effective batch of 8. We chose the learning rate from $\{10^{-5}, 2 \times 10^{-5}\}$ by training \sF\ and \sC\ on the training split and scoring them on the development split: $10^{-5}$ gives 64.1 action F1 and 41.8 intent F1, against 59.7 and 40.6 for $2 \times 10^{-5}$. Answers are decoded greedily with at most 400 new tokens, and only the first JSON object in an answer is parsed. The answers of the zero-shot model average about 83 tokens, a short reasoning followed by the JSON, and 5{,}409 of its 5{,}427 intent answers parse.

\section{Per-Seed Scores, Strata and Other Backbones}
\label{app:perseed}

Fine-tuning results move with the random seed \citep{dodge2020fine}, so Table~\ref{tab:perseed} lists the action and intent scores of every Qwen3.5 4B arm for each seed, and Table~\ref{tab:strata} splits the three-seed means by group size. The verdicts in the main text use the means of these columns.

\begin{table}[t]
\centering
\small
\setlength{\tabcolsep}{3.2pt}
\begin{tabular}{l ccc ccc}
\toprule
& \multicolumn{3}{c}{Action $F$} & \multicolumn{3}{c}{Intent $C$} \\
\cmidrule(lr){2-4} \cmidrule(lr){5-7}
Arm & 42 & 1 & 2 & 42 & 1 & 2 \\
\midrule
S-F & 60.4 & 59.1 & 59.9 & -- & -- & -- \\
S-C & -- & -- & -- & 29.1 & 30.2 & 31.0 \\
MIX & 58.3 & 56.8 & 57.6 & 30.9 & 29.4 & 31.4 \\
J-FC & 59.5 & 58.5 & 61.1 & 31.0 & 30.5 & 31.5 \\
J-CF & 59.4 & 58.6 & 60.8 & 31.8 & 30.5 & 31.5 \\
COND & 57.7 & 57.2 & 56.1 & 28.1 & 27.7 & 27.5 \\
COND-N & 58.3 & 58.5 & 59.1 & 30.8 & 30.3 & 29.3 \\
COND-S & 58.7 & 57.9 & 58.4 & 29.5 & 29.0 & 30.1 \\
\midrule
S-C-P & -- & -- & -- & 29.4 & 29.5 & 29.9 \\
J-nFC & -- & -- & -- & 29.8 & 29.2 & 30.4 \\
J-aFC & -- & -- & -- & 26.9 & 29.4 & 28.1 \\
J-sFC & -- & -- & -- & 29.8 & 30.8 & 30.7 \\
\bottomrule
\end{tabular}
\caption{Scores ($\times 100$) of each Qwen3.5 4B arm for seeds 42, 1 and 2. The content controls write donor or empty actions by design, so their action scores are not meaningful (4.4 to 4.7 for \textsc{J-sFC}, below the majority prior).}
\label{tab:perseed}
\end{table}

\begin{table*}[t]
\centering
\small
\setlength{\tabcolsep}{4.5pt}
\begin{tabular}{l cc cc cc l cc l}
\toprule
& \multicolumn{2}{c}{Action $F$} & \multicolumn{2}{c}{Intent $C$} & \multicolumn{3}{c}{Multi-member groups} & \multicolumn{3}{c}{Single people} \\
\cmidrule(lr){2-3} \cmidrule(lr){4-5} \cmidrule(lr){6-8} \cmidrule(lr){9-11}
Arm & multi & single & multi & single & $\Delta_F$ & $\Delta_C$ & verdict & $\Delta_F$ & $\Delta_C$ & verdict \\
\midrule
S-F & 56.8 & 52.1 & -- & -- & & & reference & & & reference \\
S-C & -- & -- & 31.6 & 25.0 & & & reference & & & reference \\
MIX & 54.2 & 50.6 & 32.0 & 25.5 & $-2.5$ & $+0.4$ & \textsc{Trades} & $-1.5$ & $+0.5$ & \textsc{Trades} \\
J-FC & 56.9 & 50.6 & 31.6 & 25.9 & $+0.2$ & $-0.1$ & \textsc{Trades} & $-1.5$ & $+0.9$ & \textsc{Trades} \\
J-CF & 56.0 & 52.0 & 31.4 & 26.1 & $-0.8$ & $-0.2$ & \textsc{Harms} & $-0.1$ & $+1.1$ & \textsc{Trades} \\
COND & 54.4 & 48.9 & 28.4 & 24.3 & $-2.4$ & $-3.2$ & \textsc{Harms} & $-3.2$ & $-0.8$ & \textsc{Harms} \\
COND-N & 54.8 & 51.2 & 31.7 & 24.5 & $-2.0$ & $+0.1$ & \textsc{Trades} & $-0.9$ & $-0.5$ & \textsc{Harms} \\
COND-S & 55.4 & 49.7 & 30.1 & 25.1 & $-1.4$ & $-1.5$ & \textsc{Harms} & $-2.4$ & $+0.1$ & \textsc{Trades} \\
\bottomrule
\end{tabular}
\caption{Three-seed means of Qwen3.5 4B split by group size, with the verdict rule applied inside each stratum (250 multi-member groups, 886 single people).}
\label{tab:strata}
\end{table*}

\paragraph{Seeds and subsets.}
Per seed, \jfc\ beats \sC\ on intent by 1.9, 0.3 and 0.5 points on all groups, by 2.0, $-0.8$ and $-1.4$ on multi-member groups and by 0.2, 1.1 and 1.2 on single people. On the visual subset \sC\ scores 30.5, 31.1 and 33.7 and \jfc\ adds 3.1, 1.0 and 0.8. On the five families the visual-subset gains are 0.9 for \jfc, 1.1 for \jcf\ and 1.5 for \textsc{MIX}, and on the eight intents without discussing 0.7, 1.2 and 0.9: about 60\% of the \jfc\ gain on the subset comes from discussing (0 to 8.1), since 22 of the 24 discussing groups fall into it. Inside the subset, single people gain 2.4, 2.6 and 2.0 points, for \jfc\ mostly on navigating (5.8 to 13.2) and waiting (10.7 to 20.5), and multi-member groups gain 0.0, $-0.1$ and 0.7. On the 800 groups outside the subset the three arms stay within 0.6 points of \sC.

\paragraph{Controls by stratum.}
\jfc\ is 1.6, 1.0 and 1.6 points above the prompt-matched \textsc{S-C-P} for the three seeds, 0.4 on multi-member groups and 1.1 on single people. At $2 \times 10^{-5}$ (seed 42) \jfc\ is $-0.6$ / $+2.7$ on multi-member groups and $+1.2$ / $+2.8$ on single people against the references of the same learning rate. The six-epoch \sC\ (seed 42) reaches 46.2 on the five families, 1.3 below its three-epoch score.

\begin{table}[t]
\centering
\small
\setlength{\tabcolsep}{3.5pt}
\begin{tabular}{ll rr rr}
\toprule
Backbone & Arm & $F$ & $C$ & $\Delta_F$ & $\Delta_C$ \\
\midrule
Qwen3.5 9B & zero shot & 36.1 & 17.8 & & \\
(seed 42) & S-F & 57.6 & -- & & \\
 & S-C & -- & 26.5 & & \\
 & MIX & 54.0 & 27.4 & $-3.5$ & $+1.0$ \\
 & J-FC & 56.0 & 28.0 & $-1.5$ & $+1.5$ \\
 & COND & 52.2 & 23.8 & $-5.4$ & $-2.7$ \\
\midrule
Qwen3-VL 4B & zero shot & 34.5 & 24.4 & & \\
(3 seeds) & S-F & 55.6 & -- & & \\
 & S-C & -- & 28.7 & & \\
 & MIX & 53.6 & 29.2 & $-2.0$ & $+0.5$ \\
 & J-FC & 54.8 & 27.6 & $-0.8$ & $-1.2$ \\
 & J-CF & 53.9 & 28.8 & $-1.7$ & $+0.0$ \\
 & COND & 51.4 & 27.6 & $-4.2$ & $-1.1$ \\
\bottomrule
\end{tabular}
\caption{Absolute scores on the other two backbones. Deltas are computed before rounding.}
\label{tab:backbones}
\end{table}

\paragraph{Other backbones.}
Table~\ref{tab:backbones} gives the absolute scores behind the boundary rows of Table~\ref{tab:boundaries}. The 9B model starts lower on intent than the 4B model (17.8 against 26.5 zero shot) and gains more from fine-tuning (8.7 points for \sC\ against 3.6). On 9B, \jfc\ is $-0.1$ / $+1.0$ on multi-member groups and $-1.2$ / $+0.7$ on single people. On Qwen3-VL, \textsc{MIX} is $-2.3$ / $-0.2$ on multi-member groups and $-1.4$ / $+1.2$ on single people, \jfc\ $-1.7$ / $-2.1$ and $-1.3$ / $+0.5$, \jcf\ $-1.9$ / $+0.2$ and $-2.4$ / $-0.5$, and \textsc{COND} $-3.7$ / $-1.8$ and $-3.2$ / $+0.1$.

\section{Intent Classes, Content Controls, Label Budget and Test Locations}
\label{app:perclass}

\begin{table}[t]
\centering
\small
\setlength{\tabcolsep}{2.4pt}
\begin{tabular}{l rrrrrrr}
\toprule
Arm & com. & eat. & soc. & nav. & stu. & dis. & wai. \\
\midrule
S-C & 80.4 & 64.9 & 29.8 & 6.1 & 48.3 & 0.0 & 11.3 \\
J-FC & 80.6 & 66.4 & 23.6 & 9.9 & 46.3 & 6.8 & 14.3 \\
J-CF & 80.1 & 65.3 & 25.8 & 8.8 & 48.4 & 7.0 & 14.6 \\
MIX & 81.1 & 66.5 & 24.6 & 4.9 & 48.9 & 5.0 & 13.5 \\
J-sFC & 77.3 & 63.0 & 24.9 & 7.6 & 48.9 & 9.5 & 12.2 \\
COND & 79.6 & 64.6 & 15.0 & 3.6 & 47.0 & 2.7 & 9.6 \\
\midrule
Oracle & 83.6 & 47.0 & 47.9 & 16.6 & 38.4 & 13.8 & 25.4 \\
\bottomrule
\end{tabular}
\caption{Per-class intent F1, three-seed means. Leisure is 0 for every arm. The last row is the action oracle of Table~\ref{tab:classes}.}
\label{tab:perclassarms}
\end{table}

\paragraph{Per class.}
Table~\ref{tab:perclassarms} extends Figure~\ref{fig:perclass} to the other arms. \sC\ never predicts discussing correctly in any seed. \jfc\ recovers about seven fewer socializing groups per seed, and of the 441 socializing group and seed pairs only 2 are predicted as discussing, so the drop is a loss of socializing recall rather than a shift between the two social classes. Over three seeds \jfc\ finds 3 of the 72 discussing group and seed pairs and predicts discussing for 11 pairs outside the social family, 9 of them single people; \textsc{J-sFC} finds 4. Leaving discussing out of the macro average gives 34.4 for \sC\ and 34.5 for \jfc; leaving out discussing and socializing gives 35.2 and 36.3 (\jcf\ 34.7 and 36.2, \textsc{MIX} 34.2 and 35.8). The per-class gain of \jfc\ is barely related to how much room the action oracle leaves above \sC\ (rank correlation 0.10 over the eight classes): socializing has the most room (18 points) and loses most.

\paragraph{Collapse to frequent intents.}
\sC\ predicts 7, 6 and 6 distinct intents in the three seeds, \jfc\ 7 in each, \jcf\ 8, 7 and 7, and \textsc{COND-S} 7, 9 and 8. The entropy of the predicted intent distribution is 1.54, 1.46 and 1.58 bits for \sC, 1.60, 1.63 and 1.57 for \jfc, and 1.39 to 1.40 for \textsc{COND}, which answers commuting for 738 test groups against 642 for \sC\ (seed 42). Every arm over-predicts eating (288 to 382 predicted groups against 211 gold, seed 42) and none recalls leisure.

\begin{table}[t]
\centering
\small
\setlength{\tabcolsep}{3.2pt}
\begin{tabular}{l r r rrr}
\toprule
Opus recovers it from & Pairs & S-C & J-FC & J-CF & MIX \\
\midrule
people only & 101 & 55.8 & $-3.3$ & $-0.3$ & $+1.0$ \\
people and place & 338 & 95.7 & $-1.5$ & $-0.9$ & $+0.3$ \\
place only & 247 & 89.3 & $+1.2$ & $+0.4$ & $+0.9$ \\
neither & 89 & 55.1 & $-1.1$ & $-4.5$ & $-3.0$ \\
not even the full view & 543 & 23.4 & $-0.1$ & $+0.1$ & $+0.1$ \\
\bottomrule
\end{tabular}
\caption{Recall of gold (group, intent) pairs, split by the views from which Opus ranks that intent first, three-seed means. Arm columns are differences from \sC.}
\label{tab:cues}
\end{table}

\paragraph{Where the cue lies.}
Opus also saw each test group in the full view, with only the people and with only the place. A gold intent enters a stratum only when Opus ranks it first (ties allowed) in the full view; it then counts as recoverable from a partial view when Opus also ranks it first there and does not answer that it cannot tell. All other gold intents form the last row. Table~\ref{tab:cues} gives the recall per stratum. The gain of the joint arms does not sit on the intents that only people reveal: \jfc\ recalls 3.3 points fewer of them than \sC, and 5.7 fewer on multi-member groups.

\paragraph{Content controls.}
Table~\ref{tab:content} repeats the content controls of Figure~\ref{fig:content} on every subset. \textsc{J-sFC} reaches 9.5 on discussing, the highest of all arms, so wrong actions also teach the model to use that label; without discussing, \textsc{J-sFC} is 1.0 point below \sC. On the visual subset \textsc{J-sFC} equals \jfc.

\begin{table}[t]
\centering
\small
\setlength{\tabcolsep}{2.5pt}
\begin{tabular}{ll rrrrr}
\toprule
Arm & Actions & All & Multi & Single & 5-cl. & Vis. \\
\midrule
S-C & none & 30.1 & 31.6 & 25.0 & 48.7 & 31.8 \\
J-nFC & empty lists & 29.8 & 30.9 & 24.4 & 46.9 & 31.3 \\
J-aFC & donor, same intent & 28.1 & 30.0 & 23.6 & 45.2 & 30.4 \\
J-sFC & donor, other intent & 30.4 & 31.5 & 25.0 & 47.7 & 33.4 \\
J-FC & correct & 31.0 & 31.6 & 25.9 & 48.3 & 33.4 \\
\bottomrule
\end{tabular}
\caption{Intent macro-F1 ($\times 100$, three seeds) when the action part of the joint target is emptied or replaced by the actions of a donor group, mostly from another location. All arms share the J-FC instruction and output format except S-C.}
\label{tab:content}
\end{table}

\paragraph{Label budget.}
All label budget runs use seed 42. Per draw, \sC\ scores 27.0 and 23.5 at 25\% and 29.1 and 26.0 at 50\%, while \jfc\ scores 27.1 and 28.6, then 30.0 and 28.1. With few intent labels \sC\ depends strongly on which groups were drawn, and \jfc, which still learns from the actions of every group, varies much less; in the second 25\% draw \sC\ falls to 23.1 on multi-member groups, against 30.0 for \jfc. Without discussing, the gap is 1.6, 1.7 and 1.1 points at 25\%, 50\% and 100\%. The action score of \jfc\ stays at 60.4 and 60.1 at 25\% and 59.1 at 50\%, close to \sF\ (60.4).

\begin{table}[t]
\centering
\small
\setlength{\tabcolsep}{4pt}
\begin{tabular}{l rrr}
\toprule
& same day & other day & unseen \\
Arm & (651) & (58) & (427) \\
\midrule
S-C & 28.4 & 23.0 & 31.5 \\
MIX & 27.8 & 28.6 & 31.7 \\
J-FC & 28.2 & 33.2 & 30.5 \\
J-CF & 28.9 & 30.4 & 30.6 \\
COND & 25.7 & 23.9 & 28.3 \\
AG-C & 27.2 & 30.1 & 30.9 \\
AG-FC & 29.2 & 26.5 & 29.9 \\
\midrule
S-F (action) & 52.9 & 58.5 & 54.4 \\
J-FC (action) & 53.0 & 58.2 & 54.0 \\
\bottomrule
\end{tabular}
\caption{Intent F1 by test location (three seeds; number of groups in brackets): same day and location as a training sequence, a training location on another day, or an unseen location. The last two rows give action F1.}
\label{tab:locations}
\end{table}

\paragraph{Test locations.}
Table~\ref{tab:locations} splits the test set by how close each sequence is to the training data. The gain of \jfc\ sits in the 58 groups recorded at a training location on another day. Unseen locations give higher intent scores than seen ones for most arms, probably because commuting, the easiest class, is more frequent there.

\section{Tool-Using Agent}
\label{app:agent}

\paragraph{Harness.}
The agent receives the target window and the member list, and may call up to six tools before giving its final JSON: \texttt{zoom(member)}, \texttt{pair(member\_a, member\_b)}, \texttt{frames()} for the frames two seconds before and after, \texttt{look\_around(direction)} and \texttt{overview()} for the full panorama. The system prompt is the one above with one added sentence asking for a short sentence before each call. Unlike agents that learn when to look closer through RL \citep{zheng2026deepeyes}, ours follows a fixed calling plan, so the arms differ only in what they write.

\paragraph{Trajectories.}
Training trajectories are written by a script and use only the geometry: the agent first calls \texttt{frames()}, then zooms into up to four members in order of increasing box height, then answers. The four arms share these calls and images and differ only in the text: \textsc{AG-F} and \textsc{AG-FC} write a note ``Member $k$: \ldots'' with the gold actions after each zoom, \textsc{AG-C} writes only the short sentence before each call, and \textsc{AG-CF} additionally states a first impression of the intent before its first call. The final answers follow the F, C, J-FC and J-CF formats. Training uses the supervised recipe of Appendix~\ref{app:prompts} with the loss on assistant turns only.

\paragraph{Call order.}
A first version of the trajectories called \texttt{frames()} after the zooms, so every member note was written before the agent had seen the motion. With that order (seed 42) \textsc{AG-FC} reaches only 56.3 action F1, against 58.9 with the final order and 59.5 for \jfc; by member box height its action F1 is 56.9, 53.7 and 64.3 for the small, middle and large tercile, against 56.1, 62.9 and 69.8 with the final order and 56.3, 62.4 and 69.7 for \jfc. Against \jfc, the largest per-action losses of the first version are cycling ($-10.4$), talking ($-7.5$) and listening ($-6.8$), actions that need motion or a second person. Its verdicts were $-0.2$ / $-0.9$ for \textsc{AG-FC} and $+0.5$ / $-4.3$ for \textsc{AG-CF}.

\begin{table}[t]
\centering
\small
\setlength{\tabcolsep}{3.5pt}
\begin{tabular}{l rrrrr}
\toprule
Arm & $F$ & $C$ & multi & single & 5-class \\
\midrule
AG-F & 60.0 & -- & -- & -- & -- \\
AG-C & -- & 30.9 & 32.7 & 24.6 & 48.3 \\
AG-FC & 59.7 & 30.4 & 31.0 & 25.6 & 49.4 \\
AG-CF & 58.4 & 28.8 & 30.3 & 23.4 & 44.8 \\
\bottomrule
\end{tabular}
\caption{Agent arms on Qwen3.5 4B, three seeds. The last three columns split the intent score.}
\label{tab:agent}
\end{table}

\paragraph{Results.}
Table~\ref{tab:agent} gives the agent scores. On multi-member groups \textsc{AG-FC} is $-0.6$ / $-1.7$ and \textsc{AG-CF} $-0.2$ / $-2.5$ against the single-task agents; on single people they are $-0.8$ / $+1.0$ and $-2.5$ / $-1.3$. \textsc{AG-CF} loses most, probably because its first impression is written before any evidence is seen; its intent change per seed is $-4.9$, $-2.3$ and $+0.8$. The tools help intent by themselves: for seed 42, \textsc{AG-C} reaches 31.4 against 29.1 for \sC, and \textsc{AG-F} 60.8 against 60.4 for \sF. Without training the tools do not help the 4B model: the zero-shot agent scores 33.8 / 20.5 against 34.8 / 26.5 for zero-shot direct answers, calls 2.46 tools per episode, mostly \texttt{look\_around}, and exceeds the budget in 109 episodes. The zero-shot 9B agent scores 36.3 / 22.3 against 36.1 / 17.8.

\paragraph{Teacher trajectories.}
A teacher run (seed 42) uses trajectories from Claude Opus 5.5, which ran the same harness on 3{,}210 training and development episodes. We keep the 1{,}617 episodes whose intent set matches the gold, whose member actions reach a set F1 of at least 0.5 and that stay within the tool budget (1{,}432 single, 185 multi-member). Trained on these episodes alone, the student reaches 51.0 and 24.5, against 50.5 and 24.0 for scripted trajectories on the same episodes, while calling 1.81 tools per episode instead of 2.20 and using \texttt{pair} 767 times. Mixing the teacher episodes into the full scripted set gives 59.2 and 31.9, against 58.9 and 30.8 without them, with $+1.8$ intent on single people and $-0.9$ on multi-member groups; the mixed student keeps the scripted tool use.

\paragraph{Observation swaps.}
We replace the image returned by every zoom call while keeping the tool text (seed 42). With another member of the same group, the intent of \textsc{AG-FC} changes by 0.049 on multi-member groups and that of \textsc{AG-C} by 0.023; with a grey image by 0.109 and 0.066; with a member of another group by 0.284 and 0.231. The intent F1 of \textsc{AG-FC} drops from 30.8 to 29.5 with grey images, 30.6 with another member and 25.5 with another group, and its action F1 from 58.9 to 52.3, 58.4 and 43.3. Occluding people or places in every image the agent can request gives a person share $\rho$ of 0.49 for \textsc{AG-FC} and 0.50 for \textsc{AG-C}; on multi-member groups hiding the people costs about 11 points and hiding the place about 6.

\paragraph{Forced prefixes in the agent.}
Table~\ref{tab:i2agent} compares two versions of the agent prefix test. In the first only the final answer is replaced; in the second every earlier note is replaced as well, so the forced content is the only version of that level in the context. With only the final turn replaced, multi-member groups already show the asymmetry of direct generation ($+1.2$ / $-2.0$ / $-1.8$ from actions to intent, $+0.9$ / $-0.3$ / $+0.3$ from intent to actions). Dropping the note of the final turn costs 1.2 and 1.8 points by itself, so these numbers are relative to the forced own answer.

\begin{table}[t]
\centering
\small
\setlength{\tabcolsep}{3.5pt}
\begin{tabular}{ll rrr}
\toprule
Direction & Replaced & Correct & Wrong & Empty \\
\midrule
actions $\to$ intent & final turn & $+1.7$ & $-3.1$ & $-1.9$ \\
 & all turns & $+1.9$ & $-2.4$ & $-2.3$ \\
intent $\to$ actions & final turn & $+1.8$ & $-3.4$ & $-1.9$ \\
 & all turns & $+1.4$ & $-2.5$ & $-0.5$ \\
\bottomrule
\end{tabular}
\caption{Forced prefixes in the agent (seed 42), relative to the forced own answer. With only the final turn replaced, the first impression of \textsc{AG-CF} still states an intent, and an empty intent then costs the actions 1.9 points.}
\label{tab:i2agent}
\end{table}

\section{GRPO Training}
\label{app:grpo}

\paragraph{Objective.}
Starting from \jfc\ (seed 42), we sample $G = 8$ answers $o_1,\dots,o_G$ per prompt at temperature 1, score each with a reward $r_i$ and use the group-centred advantage $\hat{A}_i = r_i - \frac{1}{G}\sum_j r_j$ \citep{shao2024deepseekmath,guo2025deepseek}. Groups in which all rewards are equal are dropped. For a chunk of prompts with sampled answers $\mathcal{O}$ the loss is
\begin{equation}
\begin{split}
\mathcal{L}_{\text{GRPO}} = - \frac{1}{N} \sum_{o_i \in \mathcal{O}} \sum_{t} \Bigl[ &\hat{A}_i \log \pi_\theta(o_{i,t} \mid h_{i,t}) \\
&- \beta\, \mathrm{KL}_{i,t} \Bigr],
\end{split}
\end{equation}
where $N$ is the number of generated tokens in the chunk, $h_{i,t} = (X, q, o_{i,<t})$ is the context of token $t$, $\mathrm{KL}_{i,t}$ is the k3 estimate of the divergence from the frozen \jfc\ model and $\beta = 0.02$. Answers that do not end within 400 tokens are left out of the loss. Each chunk of 16 prompts gives one update, so the importance ratio is 1. We use 1{,}024 prompts for two rounds, learning rate $2 \times 10^{-6}$ with 5\% warmup and a cosine schedule.

\paragraph{Rewards.}
Let $R_C$ be the set F1 between predicted and gold intents and $R_F$ the mean set F1 over members (difficulty-3 actions are ignored). With a format term $r_{\text{fmt}} \in \{0, 1\}$ for a parsable answer that holds both fields, the arms are
\begin{align}
r_{\textsc{RC}} &= R_C + r_{\text{fmt}}, \qquad r_{\textsc{RF}} = R_F + r_{\text{fmt}}, \\
r_{\textsc{RFC}} &= R_C + 0.5\, R_F + r_{\text{fmt}},
\end{align}
and \textsc{RFC-gated} adds $0.5\, R_F$ only when $R_C \ge 0.5$.

\paragraph{Room for RL.}
Before training we checked that the rewards still vary. On 256 training prompts the greedy \jfc\ answers already score $R_C = 0.885$ and $R_F = 0.935$, best-of-8 sampling reaches 0.970 and 0.972, and the eight samples differ in reward for 52\% and 61\% of the prompts, enough for group-centred advantages.

\begin{table}[t]
\centering
\small
\setlength{\tabcolsep}{4pt}
\begin{tabular}{l rr rr}
\toprule
& \multicolumn{2}{c}{train+dev prompts} & \multicolumn{2}{c}{dev prompts} \\
\cmidrule(lr){2-3} \cmidrule(lr){4-5}
Arm & $\Delta_F$ & $\Delta_C$ & $\Delta_F$ & $\Delta_C$ \\
\midrule
RC & $-0.1$ & $+0.8$ & $+0.1$ & $-1.8$ \\
RF & $-1.1$ & $0.0$ & $-2.5$ & $+0.1$ \\
RFC & $-1.0$ & $+0.1$ & $-2.0$ & $-3.3$ \\
RFC-gated & $-1.0$ & $-0.4$ & $+0.3$ & $-2.7$ \\
\bottomrule
\end{tabular}
\caption{GRPO changes relative to the starting model. Left: start from \jfc\ trained on train and development data, RL on the same prompts (two seeds). Right: start from \jfc\ trained on the training split only (57.4 / 29.6), RL on development prompts (seed 42).}
\label{tab:grpo}
\end{table}

\begin{table}[t]
\centering
\footnotesize
\setlength{\tabcolsep}{2.5pt}
\begin{tabular}{l ll c}
\toprule
Run & $R_C$ & $R_F$ & varying \\
\midrule
RC, seen & 0.867 $\to$ 0.889 & & 0.43 $\to$ 0.30 \\
RF, seen & & 0.871 $\to$ 0.916 & 0.52 $\to$ 0.39 \\
RFC, seen & 0.871 $\to$ 0.884 & 0.874 $\to$ 0.905 & 0.69 $\to$ 0.59 \\
RC, dev & 0.684 $\to$ 0.744 & & 0.53 $\to$ 0.36 \\
RF, dev & & 0.717 $\to$ 0.765 & 0.70 $\to$ 0.48 \\
RFC, dev & 0.677 $\to$ 0.738 & 0.715 $\to$ 0.754 & 0.82 $\to$ 0.64 \\
RC, agent & 0.830 $\to$ 0.861 & & 0.47 $\to$ 0.28 \\
RFC, agent & 0.839 $\to$ 0.863 & 0.827 $\to$ 0.886 & 0.70 $\to$ 0.41 \\
\bottomrule
\end{tabular}
\caption{Training rewards, first 16 steps $\to$ last 16 steps (seed 42), and the share of prompt groups whose samples differ in reward. ``Seen'': RL prompts from the supervised training data; ``dev'': unseen development prompts; ``agent'': GRPO from \textsc{AG-FC}.}
\label{tab:dynamics}
\end{table}

\paragraph{Results.}
Table~\ref{tab:grpo} shows that \textsc{RF} and \textsc{RFC} lower action F1 in both settings, and that an intent reward helps only when the RL prompts were already seen in supervised training. On seen prompts the intent gain of \textsc{RC} is 1.3 points for seed 42 and 0.3 for seed 1, the five-family deltas are $+0.9$, $+0.4$, $+0.6$ and $+0.6$ for the four arms, and \textsc{RF} and \textsc{RFC} lose 1.5 and 1.6 action points on multi-member groups. With unseen development prompts the five-family deltas are $-4.4$, $-1.4$, $-5.1$ and $-4.0$. Table~\ref{tab:dynamics} shows that the rewarded quantity rises in every run while the share of prompts with varying rewards falls, and the KL per token stays at or below 0.01. On held-out development groups the reward falls instead (0.82 to 0.78 for \textsc{RC}, 0.80 to 0.77 for \textsc{RF}), and so do the test scores. The number of actions per member moves only from 1.23 to 1.19 under \textsc{RF} (seed 42), so the reward is not gamed by empty answers.

\paragraph{GRPO on the agent.}
For the agent (seed 42), GRPO from \textsc{AG-FC} with 768 prompts gives $+0.4$ / $-0.8$ for \textsc{RC}, $+0.4$ / $-0.3$ for \textsc{RFC} and $-0.8$ / $-0.5$ for a variant that adds 0.1 to a correct intent when the agent called a tool. The starting agent calls a tool in every episode, so this variant only rewards correct intents more. The tool calls per episode stay at 2.19.

\paragraph{Mechanisms after GRPO.}
GRPO leaves the mechanisms of \S\ref{sec:analysis} where they were (seed 42). The person share $\rho$ of the four direct arms is 0.48, 0.41, 0.47 and 0.46 against 0.46 at the start, and 0.55, 0.52 and 0.49 for the agent arms against 0.49. Hiding one member changes their intent by 0.192 to 0.201, with marked-member hit rates of 0.527 to 0.546, and swapping a zoom image for another group's member changes the agent intent on multi-member groups by 0.269 to 0.292.

\section{Mechanism Details}
\label{app:mechanism}

\begin{table*}[t]
\centering
\small
\setlength{\tabcolsep}{5pt}
\begin{tabular}{l rrrrrrrr}
\toprule
Arm & Intent change & Gold-intent drop & Others & Pairs & Marked & Chance & Strong & Weak \\
\midrule
S-F & -- & -- & 0.216 & -- & -- & -- & -- & -- \\
S-C & 0.163 & 0.056 & -- & 162 & 0.545 & 0.471 & 0.190 & 0.159 \\
J-FC & 0.194 & 0.067 & 0.188 & 179 & 0.555 & 0.477 & 0.233 & 0.197 \\
J-CF & 0.177 & 0.050 & 0.229 & 151 & 0.535 & 0.478 & 0.202 & 0.159 \\
MIX & 0.174 & 0.054 & -- & 170 & 0.567 & 0.480 & 0.186 & 0.171 \\
COND & 0.246 & 0.103 & -- & 212 & 0.470 & 0.469 & 0.232 & 0.258 \\
\midrule
AG-F & -- & -- & 0.234 & -- & -- & -- & -- & -- \\
AG-C & 0.161 & 0.048 & -- & 160 & 0.509 & 0.471 & 0.179 & 0.155 \\
AG-FC & 0.214 & 0.061 & 0.243 & 203 & 0.471 & 0.475 & 0.212 & 0.212 \\
\bottomrule
\end{tabular}
\caption{Hiding one member (seed 42, 1{,}926 member and sample pairs). Intent change: change of the predicted intent set. Gold-intent drop: loss of set F1 against the gold intent. Pairs: (group, intent) pairs with a marked member that enter the hit rate. Marked: share of the marked member among the members whose removal deletes a correct intent. Strong and weak: intent change when a member rated as strong or weak evidence is hidden.}
\label{tab:i4}
\end{table*}

\begin{figure*}[t]
\centering
\includegraphics[width=\textwidth]{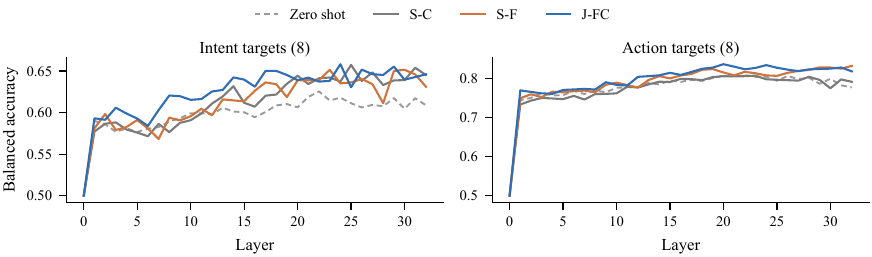}
\caption{Linear probe balanced accuracy per layer, averaged over eight intent targets (left) and eight action targets (right). Seed 42; every model reads the J-FC prompt.}
\label{fig:probes}
\end{figure*}

\begin{figure*}[t]
\centering
\includegraphics[width=\textwidth]{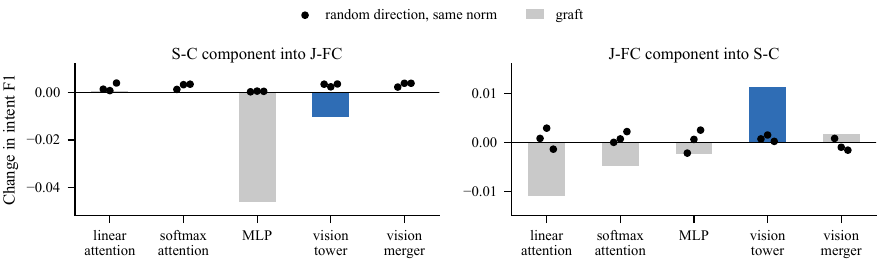}
\caption{Component grafting (seed 42). Bars: change in intent F1 when one component group of the donor replaces that of the base model; the vision tower is shown in blue. Dots: three random perturbations of the same tensors with the same norm.}
\label{fig:grafting}
\end{figure*}

\paragraph{Occlusion.}
The ``no people'' view covers each member box grown by 25\% (wider for cyclists and people pushing carts); the ``no place'' view keeps each box grown by 5\% and covers the rest. Both are filled with grey; a blurred fill is the robustness variant. Table~\ref{tab:i1} lists $\rho$ per seed. For seed 42, \sC\ scores 29.1 on the full view, 20.4 with only the people and 25.5 with only the place; \textsc{COND} scores 28.1, 22.5 and 14.6. The masks act on actions as expected: \sF\ keeps 49.4 action F1 with only the people and drops to 24.0 with only the place.

\begin{table}[t]
\centering
\small
\setlength{\tabcolsep}{4pt}
\begin{tabular}{l ccc c}
\toprule
Arm & 42 & 1 & 2 & Mean \\
\midrule
zero shot & & 0.42 & 0.42 & 0.42 \\
S-C & 0.29 & 0.47 & 0.43 & 0.40 \\
MIX & 0.60 & 0.44 & 0.47 & 0.50 \\
J-FC & 0.46 & 0.45 & 0.48 & 0.46 \\
J-CF & 0.49 & 0.44 & 0.60 & 0.51 \\
COND & 0.71 & 0.43 & 0.64 & 0.59 \\
\midrule
S-C, blur & 0.34 & 0.58 & 0.62 & 0.51 \\
J-FC, blur & 0.52 & 0.51 & 0.55 & 0.53 \\
\bottomrule
\end{tabular}
\caption{Person share $\rho$ of the occlusion loss per seed.}
\label{tab:i1}
\end{table}

\paragraph{Forced prefixes per seed.}
Table~\ref{tab:i2seeds} lists the prefix effects of direct generation per seed. The empty prefix of seed 2 makes \jfc\ answer commuting for 5{,}403 of the 5{,}427 samples. For seed 42, \jfc\ with gold actions written reaches 34.4 on multi-member groups, 27.4 on single people and 51.6 on the five families.

\begin{table}[t]
\centering
\small
\setlength{\tabcolsep}{3.5pt}
\begin{tabular}{l l rrr}
\toprule
Direction & Prefix & 42 & 1 & 2 \\
\midrule
actions $\to$ intent & correct & $+2.9$ & $+1.3$ & $+2.7$ \\
(\jfc) & wrong & $-2.4$ & $-4.0$ & $-3.2$ \\
 & empty & $-4.1$ & $-3.7$ & $-22.6$ \\
\midrule
intent $\to$ actions & correct & $+1.0$ & $+1.6$ & $+1.3$ \\
(\jcf) & wrong & $-3.4$ & $-3.4$ & $-3.0$ \\
 & empty & $0.0$ & $-0.6$ & $-0.8$ \\
\bottomrule
\end{tabular}
\caption{Forced prefixes in direct generation per seed, change of the decoded side's macro-F1 ($\times 100$).}
\label{tab:i2seeds}
\end{table}

\paragraph{Member masking.}
Table~\ref{tab:i4} adds the remaining arms. ``Others'' is the change in the predicted actions of the members that stay visible. The single-task \sF\ shows this coupling as well (0.216), so it does not come from writing both levels in one answer. \textsc{COND} reacts most strongly to a single member but chooses members at chance level and reacts more to weak than to strong members. On the 738 pairs whose member box overlaps little with the others, hiding a member changes the intent of \jfc\ by 0.228 and that of \sC\ by 0.168, with hit rates of 0.607 and 0.631.

\paragraph{Probes.}
We sample 3{,}000 test samples once, reduce each layer's hidden state to 256 dimensions with PCA and fit an L2 logistic regression per target in five folds split by sequence. The intent targets are the eight headline intents; the action targets are the eight most frequent actions, each true when some member performs it. The zero-shot model already reaches 66.1 balanced accuracy on intents and 81.8 on actions at its best layer, well above a probe on the location alone (56.7 and 57.0) and above a permutation baseline that swaps the labels of whole groups of the same size (95th percentile over 50 permutations of the best-layer score: 56.2 and 56.7). Two targets stay at the location level: on leisure no model beats the location probe (65.1), and on navigating the zero-shot model and \textsc{COND} do not (61.3). Table~\ref{tab:i5} gives the gains over the zero-shot model, and Figure~\ref{fig:probes} the curves over layers: on intent the trained models pull away from the zero-shot model mainly in the upper half of the network, while on actions \sC\ stays close to it at every depth.

\begin{table}[h]
\centering
\small
\setlength{\tabcolsep}{3.5pt}
\begin{tabular}{l l rr rr}
\toprule
& & \multicolumn{2}{c}{Intent} & \multicolumn{2}{c}{Action} \\
\cmidrule(lr){3-4} \cmidrule(lr){5-6}
Arm & Prompt & best & late & best & late \\
\midrule
S-F & J-FC & 1.1 & 2.6 & 2.7 & 2.0 \\
S-C & J-FC & 2.0 & 2.7 & 0.5 & 0.1 \\
MIX & J-FC & 2.0 & 3.1 & 2.6 & 2.5 \\
J-FC & J-FC & 1.8 & 3.3 & 2.6 & 2.7 \\
J-CF & J-FC & 2.5 & 3.4 & 3.3 & 2.9 \\
COND & J-FC & 0.6 & 1.4 & 1.7 & 0.9 \\
\midrule
S-F & F & 3.4 & 3.6 & 1.8 & 1.8 \\
S-C & C & 2.3 & 1.7 & 0.6 & 0.2 \\
\bottomrule
\end{tabular}
\caption{Probe gain over the zero-shot model (balanced accuracy $\times 100$, seed 42) at the best layer and averaged over layers 16 to 32. The last two rows feed each single-task model its own prompt and compare with the zero-shot model under the same prompt.}
\label{tab:i5}
\end{table}

\paragraph{Grafting.}
The five component groups together hold 657 tensors; embeddings and layer norms stay with the base model, and each grafted model is tested with the base model's prompt. Figure~\ref{fig:grafting} shows every graft with its random controls. Measured against the mean of the random controls instead of the base model, the vision tower moves intent F1 by $-1.37$ and $+1.05$ points. Grafting the MLPs of \sC\ into \jfc\ breaks the action field: the model keeps the joint format but writes intent labels as member actions, action F1 falls from 59.5 to 6.3 and intent F1 by 4.6, so this graft cannot locate the gain. The language-side attention layers of \jfc\ lower the intent score of \sC\ when moved alone ($-1.1$ for linear attention, $-0.5$ for softmax attention). Grafting the vision tower of \sC\ also lowers the action F1 of \jfc\ by 1.6 points, and the vision tower of \jfc\ raises the intent of \sC\ by 1.7 points on multi-member groups and lowers it by 0.6 on single people. The gap between \jfc\ and \sC\ that grafting works on is 1.9 points for seed 42, the largest of the three seeds (0.3 and 0.5 for the others).

\end{document}